\documentclass{research}

\usepackage{amsmath,amsfonts,bm}

\def\eqref#1{equation~\ref{#1}}
\def\Eqref#1{Equation~\ref{#1}}

\def\1{\bm{1}}

\def\vtheta{{\bm{\theta}}}
\def\va{{\bm{a}}}

\def\vc{{\bm{c}}}

\def\vh{{\bm{h}}}

\def\vo{{\bm{o}}}

\def\vs{{\bm{s}}}

\def\vu{{\bm{u}}}
\def\vv{{\bm{v}}}

\def\vx{{\bm{x}}}

\def\vz{{\bm{z}}}

\DeclareMathAlphabet{\mathsfit}{\encodingdefault}{\sfdefault}{m}{sl}
\SetMathAlphabet{\mathsfit}{bold}{\encodingdefault}{\sfdefault}{bx}{n}

\newcommand{\E}{\mathbb{E}}

\newcommand{\R}{\mathbb{R}}

\newcommand{\KL}{D_{\mathrm{KL}}}

\DeclareMathOperator*{\argmin}{arg\,min}

\newcommand{\vphi}{\boldsymbol{\phi}}

\usepackage{hyperref}
\usepackage{url}

\usepackage{amsmath}
\usepackage{amssymb}

\usepackage{etoc}
\usepackage{titletoc}

\usepackage{hyperref}

\usepackage{booktabs}
\usepackage[table]{xcolor}
\usepackage{multirow}

\usepackage{array}

\usepackage{amsthm}

\usepackage{enumitem}

\usepackage{bm}
\usepackage{tcolorbox}
\tcbuselibrary{skins}
\usepackage{cleveref}

\Crefname{equation}{Equation}{Equations}
\crefname{equation}{Eq.}{Eqs.}
\Crefname{figure}{Figure}{Figures}
\crefname{figure}{Fig.}{Figs.}
\Crefname{table}{Table}{Tables}
\crefname{table}{Table}{Tables}
\Crefname{section}{Section}{Sections}
\crefname{section}{Section}{Sections}
\Crefname{appendix}{Appendix}{Appendices}
\crefname{appendix}{Appendix}{Appendices}
\AddToHook{cmd/appendix/before}{%
    \crefalias{section}{appendix}%
    \crefalias{subsection}{appendix}
}

\definecolor{sand}{RGB}{248,245,235}  
\definecolor{sandborder}{RGB}{220,215,200}
\definecolor{darkblue}{RGB}{40,70,140}   
\definecolor{parchment}{RGB}{252,249,240}
\definecolor{parchmentborder}{RGB}{222,215,195}
\definecolor{sage}{RGB}{244,248,244}
\definecolor{sageborder}{RGB}{200,210,200}
\definecolor{lavender}{RGB}{247,246,250}
\definecolor{lavenderborder}{RGB}{210,205,225}
\definecolor{riskbg}{RGB}{255,246,246}
\definecolor{riskborder}{RGB}{205,105,105}

\definecolor{LyapuCol}{RGB}{235,241,247}   
\definecolor{Satisfied}{RGB}{226,239,218}  
\definecolor{Violated}{RGB}{249,220,220}   
\definecolor{efficiencyred}{RGB}{187,94,96}
\definecolor{Gray}{gray}{0.9}

\newcolumntype{Y}{>{\columncolor{LyapuCol}}r}

\newcommand{\satoff}[1]{0}

\usepackage{float}
\floatstyle{ruled}
\newfloat{algorithm}{tbp}{loa}
\floatname{algorithm}{Algorithm}
\usepackage{capt-of}
\usepackage{url}
\usepackage{graphicx}
\usepackage{colortbl}
\usepackage{wrapfig2}
\usepackage{svg}
\newenvironment{vsfalgorithmic}{%
  \begingroup
  \footnotesize
  \raggedright
  \setlength{\parindent}{0pt}%
  \setlength{\parskip}{0.2em}%
}{%
  \par\endgroup
}
\newcommand{\VSFInput}{\par\noindent\textbf{Input:}~}
\newcommand{\VSFOutput}{\par\noindent\textbf{Output:}~}
\newcommand{\VSFState}{\par\noindent}
\newcommand{\VSFWhile}[1]{%
  \par\noindent\textbf{while} #1 \textbf{do}\par
  \begingroup\addtolength{\leftskip}{1em}%
}
\newcommand{\VSFEndWhile}{%
  \par\endgroup\noindent\textbf{end while}%
}
\newcommand{\VSFFor}[1]{%
  \par\noindent\textbf{for} #1 \textbf{do}\par
  \begingroup\addtolength{\leftskip}{1em}%
}
\newcommand{\VSFEndFor}{%
  \par\endgroup\noindent\textbf{end for}%
}

\usepackage{bm}
\usepackage{tcolorbox}
\tcbuselibrary{skins}

\definecolor{resultDPBg}{RGB}{242,247,255}
\definecolor{resultRosslerBg}{RGB}{242,250,244}
\definecolor{resultLorenzBg}{RGB}{255,247,235}
\definecolor{resultRuntimeBg}{RGB}{245,240,252}
\newlength{\resultHeaderGap}
\newtheorem{vsftheorem}{Theorem}[section]
\newtheorem{vsfcorollary}[vsftheorem]{Corollary}
\newtheorem{vsfproposition}[vsftheorem]{Proposition}
\numberwithin{equation}{section}
\providecommand{\resultpm}[2]{\ensuremath{#1 \mathbin{\pm} #2}}

\providecommand{\fullbestresultpm}[2]{\ensuremath{\boldsymbol{#1 \mathbin{\pm} #2}}}

\providecommand{\DoublePendulumChaoticInstance}{02}

\title{Variational Streaming Flow: Probabilistic Forecasting in Physical Time}

\author[1,2]{Hans Hao-Hsun Hsu}
\author[1,5]{Minseon Gwak}
\author[3,4]{Soon Hoe Lim\textsuperscript{*}}
\author[2]{\\Pan Li}
\author[1,5]{N. Benjamin Erichson}

\affiliation[1]{Lawrence Berkeley National Lab}
\affiliation[2]{Georgia Institute of Technology}
\affiliation[3]{KTH Royal Institute of Technology}
\affiliation[4]{Nordita}
\affiliation[5]{International Computer Science Institute}
\correspondence{\vspace{-1em}\email{shlim@kth.se}; \email{panli@gatech.edu}; 
and \email{erichson@lbl.gov}}

\metadata[Code]{\url{https://github.com/hans66hsu/VSF-forecasting}}

\abstract{
\vspace{-0.5em}
Probabilistic forecasting is important for predicting complex dynamical systems because intrinsic randomness and incomplete observations can cause the same observed state to evolve into multiple plausible futures. While flow matching is a flexible approach for probabilistic forecasting, it is computationally expensive. Streaming flow (SF) reformulates this approach to model temporal evolution efficiently by learning a continuous velocity field directly in physical time. However, SF learns a deterministic velocity field. Thus, it provides only a single future trajectory for a given fixed initial state and observation history. To overcome this limitation, we introduce Variational Streaming Flow (VSF). Our approach learns a latent distribution that is conditioned on the dynamics of interest. In turn, this enables probabilistic forecasting. Importantly, we retain the computational efficiency of SF by generating in physical time. Across deterministic and stochastic dynamical systems, VSF demonstrates superior predictive accuracy and distributional fidelity. We demonstrate the advantage for both long-horizon rollouts exceeding 1,000 steps, and settings with bifurcating dynamics. Moreover, VSF can be integrated into existing Joint-Embedding Predictive Architecture (JEPA)-based world models as a plug-and-play predictor to improve temporal dynamics and goal-directed success rate in navigation, motion planning, and manipulation.}

\begin{document}

\maketitle

\begingroup
\renewcommand{\thefootnote}{*}
\footnotetext{Work done while visiting ICSI/UC Berkeley.}
\endgroup

\begin{figure}[H]
\vspace{-0.5cm}
  \centering
  \includegraphics[width=0.95\linewidth]{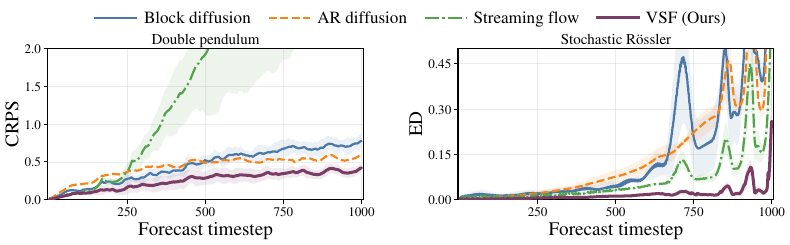}
  \vspace{-0.8em}
\caption{Our variational streaming flow approach obtains lower forecasting error over long rollouts than standard streaming flow. Left: continuous ranked probability score (CRPS) on the deterministic double pendulum. Right: energy distance (ED) on the partially observed R\"ossler system.}
  \label{fig:main-rollout-error-curves}
  \vspace{-0.15cm}
\end{figure}

\section{Introduction}

Forecasting dynamical systems over long horizons remains challenging. Even when the dynamics are deterministic, small measurement or prediction errors can compound over a long rollout~\citep{zhang2025zero}. Predicting the entire future trajectory at once can avoid this recursive error accumulation, but this becomes an increasingly difficult learning task as the forecast horizon grows. Many real-world systems are also stochastic, which means that there can be many different future trajectories for the same sequence of observations~\citep{pfister2026stocbench}. For example, in weather forecasting~\citep{ feng2013expansion,sandu2019impacts, price2025probabilistic}, unresolved atmospheric processes can lead to different storm tracks from similar observations. In these settings, predicting just a single trajectory is insufficient for forecasting or planning. This motivates probabilistic forecasts that represent a distribution over possible futures~\citep{chua2018deep,porcher2026flow}.

Flow matching provides a flexible way to learn probabilistic forecasts over entire future trajectories~\citep{lipman2024flow,kollovieh2025tsflow}. However, generating a forecast requires integrating a learned vector field along an auxiliary flow time that is separate from the physical evolution of the system. Streaming and physical-time formulations instead evolve the predicted state directly through physical time, making long-horizon forecasting more efficient~\citep{jiang2025streaming,liu2026odeworld}. However, a standard non-latent physical-time model with a deterministic velocity field produces only one future from a fixed observed state.

To address this limitation, we introduce \emph{Variational Streaming Flow} (VSF), a probabilistic extension of physical-time flow forecasting. VSF augments the physical-time velocity field with a latent variable that is sampled once for each forecast and kept fixed throughout the rollout. This preserves the computational efficiency of streaming flow (SF) while improving long-horizon forecasting accuracy, as shown in \cref{fig:main-rollout-error-curves}. Different latent samples can also generate different future trajectories from the same observed state, allowing VSF to represent multiple possible futures.
During training, we use an encoder to infer the latent variable from the observed future transition and regularize its distribution toward a fixed prior~\citep{guo2025variational,eijkelboom2024variational}. At inference, we only need to sample a latent from the prior and keep it fixed throughout the rollout.

\begin{figure}[!t]
  \centering
  \includegraphics[
    width=0.95\linewidth,
    trim={0bp 10bp 0bp 12bp},
    clip
  ]{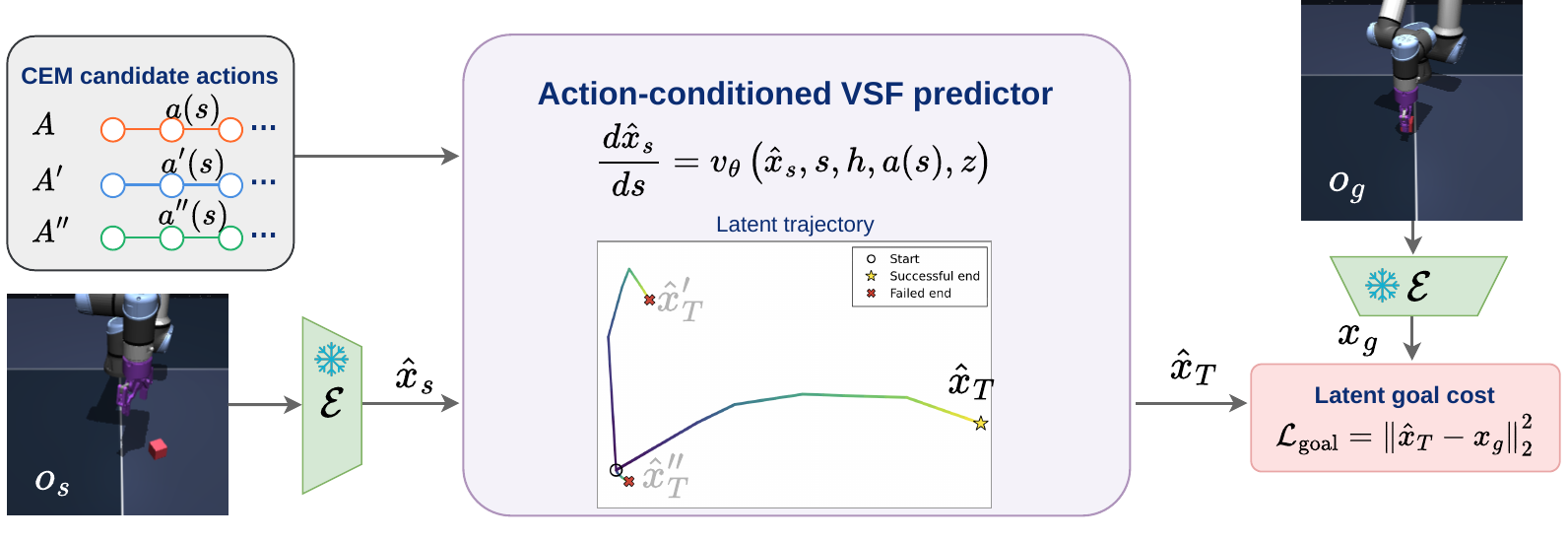}
  \vspace{-0.8em}
\caption{
VSF can serve as a predictor in a JEPA-based world model. We replace the deterministic autoregressive predictor with an action-conditioned VSF predictor while keeping the JEPA encoder and CEM planner fixed. The encoder maps the current and goal observations into latent space. For each action sequence, VSF samples multiple latents and rolls out future trajectories. CEM compares the predicted terminal states with the goal representation and uses these costs to select a plan.
}
  \label{fig:world-model-predictor}
\vspace{-1em}
\end{figure}

We evaluate VSF on deterministic and partially observed stochastic dynamical systems, including chaotic systems and settings with branching futures, testing both predictive accuracy and distributional fidelity. On deterministic systems, VSF improves long-horizon predictive accuracy over DeepAR~\citep{salinas2020deepar}, diffusion- and flow-based forecasters, SF~\citep{jiang2025streaming}, and Chronos-2~\citep{ansari2025chronos}. On stochastic R\"ossler and Lorenz systems, VSF more accurately recovers the conditional distribution of long-horizon futures. We further use VSF as the predictor in a JEPA-based world model~\citep{maes2026leworldmodel}; see \cref{fig:world-model-predictor}. Across four robotics tasks, VSF improves average goal-directed planning success by $7.34\%$ over LeWM's deterministic predictor while retaining one velocity-field evaluation per imagined transition.

Our main contributions are as follows.

\begin{itemize}[leftmargin=*]

\item We introduce \emph{Variational Streaming Flow}, a variational latent
variable formulation of physical-time flow modeling in which one latent draw
conditions a full forecast rollout. It is motivated by a limitation of
non-latent physical-time flows near bifurcations, where small differences in
state can make the learned dynamics highly sensitive to perturbations
(see App.~\ref{app_motivate}).

\item We develop a variational learning framework for latent-conditioned
physical-time velocities (see \cref{sec:method}). We derive a
conditional-velocity ELBO and introduce an energy-based reconstruction over
multiple latent samples. Under posterior--prior alignment, we show that the
resulting reconstruction controls the energy distance between predicted and
true velocity distributions.

\item We evaluate VSF on deterministic and stochastic forecasting and in an
action-conditioned world model. VSF improves long-horizon predictive accuracy
and distributional fidelity, while its predictions improve goal-directed
planning and retain the efficiency of physical-time forecasting.

\end{itemize}

\section{Related Work}
\label{sec:related-work}


\textbf{Probabilistic and generative forecasting.} Probabilistic time-series forecasting has traditionally modeled future uncertainty through autoregressive conditional distributions—using either deterministic states like DeepAR~\citep{salinas2020deepar} or stochastic latent variables like the Variational RNN~\citep{chung2015recurrent}. CLOVER~\citep{olivares2015clover} further refines traditional parametric forecasting by reparameterizing the coherent learning objective to better optimize distribution matching. TimeGrad instead replaces these parametric one-step distributions with an autoregressive diffusion model~\citep{rasul2021timegrad}. More recently, pretrained time-series models have enabled forecasting across datasets without task-specific training. Lag-Llama provides a probabilistic decoder-only model
~\citep{rasul2024lagllama}, while Chronos models scaled and quantized
time-series values with language-model architectures~\citep{ansari2024chronos}. Other foundation models instead use direct multi-horizon prediction, including TimesFM~\citep{das2024timesfm}, Moirai~\citep{woo2024moirai}, and the multivariate and covariate-conditioned Chronos-2~\citep{ansari2025chronos}. Such models can
also provide strong zero-shot forecasts for chaotic systems
~\citep{zhang2025zero}, but they do not explicitly parameterize the
underlying evolution as a physical-time dynamical system~\citep{karaouli2025foundational}. A complementary line of work uses diffusion and flow matching to model conditional future distributions~\citep{erichson2025flex}. Rolling Diffusion adapts the diffusion process across forecast lead time~\citep{ruhe2024rolling}, TSFlow introduces Gaussian-process priors for conditional flow matching~\citep{kollovieh2025tsflow}, and flow-based forecasting has been improved through informed probability paths~\citep{lim2025paths}, autoregressive conditional flows
~\citep{elgazzar2025flowtime}, and Bayesian-filtering-inspired transport between successive predictive distributions~\citep{huang2026sequential}. 

\textbf{Streaming and physical-time flows.} Another line of work removes the distinction between generative flow time and temporal evolution. Streaming Flow Policy treats an action trajectory itself as the flow trajectory, enabling actions to be generated and executed incrementally~\citep{jiang2025streaming}. For sequential data, \citet{lim2026flow} show that empirical flow matching can be interpreted as a memory-augmented continuous-time dynamical system whose velocity field is constructed from historical transitions. Most directly, ODEWorld introduces Physical-Time Flow, which learns a latent velocity field whose integration directly follows physical time~\citep{liu2026odeworld}.

\textbf{Stochastic world models and probabilistic planning.} Predictive uncertainty has long played a role in model-based control~\citep{watter2015embed,ha2018world,hafner2025mastering}. PETS combines probabilistic dynamics ensembles with trajectory sampling
~\citep{chua2018deep}, while PlaNet learns latent dynamics with both
deterministic and stochastic components and plans through imagined
trajectories~\citep{hafner2019planet}. Recent visual world models instead often predict directly in learned feature spaces. DINO-WM predicts future pretrained visual features under candidate actions and uses these predictions for test-time planning~\citep{zhou2025dinowm}, while LeWorldModel learns an end-to-end joint-embedding predictor for efficient latent-space planning~\citep{maes2026leworldmodel}. FlowWM explicitly addresses multimodal future uncertainty by applying flow matching in pretrained feature space~\citep{porcher2026flow}.

{\bf Positioning.}  Streaming Flow (SF) formulates sequence generation directly in physical time
~\citep{jiang2025streaming}, but its standard construction represents residual
stochasticity through the evolving physical state; an auxiliary-latent
extension is also possible. Variational flow matching provides a mechanism for
representing multimodal velocity fields by inferring latent variables from
observed target dynamics~\citep{guo2025variational,eijkelboom2024variational}.
We bring these ideas together in \emph{Variational Streaming Flow (VSF)}, a
conditional latent variable formulation of physical-time flow modeling for
probabilistic forecasting. During training, a variational encoder associates
observed future dynamics with latent codes, while at inference a latent drawn
from the prior conditions a complete physical-time rollout from the observed
state. Thus, we view VSF not as the first latent SF model, but as a variational
latent variable formulation of SF for probabilistic forecasting.
Table~\ref{tab:vsf_positioning} summarizes this positioning.

This combination also raises a question specific to physical-time forecasting:
why is it useful to represent future-specific information outside the evolving
physical state? We illustrate one such mechanism through a simple
bifurcation example in App.~\ref{app_motivate}.

\begin{table}[!t]
\centering
\caption{
Comparison of VSF with closely related flow-based formulations.
``Physical time'' means that flow progression is identified with the
temporal evolution of the modeled sequence, rather than an artificial
generative-time coordinate.
``History-conditioned'' denotes explicit conditioning on an observed
history.
``Variational'' denotes the use of a training-time recognition model for
latent inference.
``Future-informed latent'' indicates that the latent representation is
inferred \emph{during training} from the realized target dynamics, rather than
introduced solely as an auxiliary stochastic coordinate. 
}
\label{tab:vsf_positioning}
\small
\vspace{0.2cm}
\setlength{\tabcolsep}{5pt}
\begin{tabular}{lccccc}
\toprule
Method
& Physical time
& History-cond.
& Latent
& Variational
& Future-informed latent \\
\midrule
FM/Rectified Flow
& \xmark
& \xmark
& \xmark
& \xmark
& \xmark \\

Variational Rectified FM
& \xmark
& \xmark
& \cmark
& \cmark
& \cmark \\

SFP
& \cmark
& \cmark
& \xmark
& \xmark
& \xmark \\

SFP + auxiliary latent
& \cmark
& \cmark
& \cmark
& \xmark
& \xmark \\

\textbf{VSF (ours)}
& \cmark
& \cmark
& \cmark
& \cmark
& \cmark \\
\bottomrule
\end{tabular}
\vspace{0.3cm}
\end{table}

\section{Problem Formulation and Preliminaries}
\label{sec:preliminaries}

This section introduces the forecasting problem and reviews the physical-time formulation of SF. 

\textbf{Forecasting problem.}
We consider forecasting from an observed history
$\vx_{a:m}=(\vx_a,\ldots,\vx_m)$, where
$\vx_k\in\mathcal X\subseteq\R^d$ denotes the system state at the $k$th
observation time. Our goal is to learn a conditional distribution over future
trajectories,
\begin{equation}
    \widehat{\vx}_{m+1:T}
    \sim
    Q_{\vtheta}(\cdot\mid\vx_{a:m}),
    \label{eq:forecasting_problem}
\end{equation}
where $Q_{\vtheta}$ is the learned forecasting model. When the dynamics are
fully observed and deterministic, this distribution contains a single future.
With partial observations or stochastic dynamics, the same observed history
may instead be consistent with 
different future trajectories.

\textbf{Streaming Flow in physical time.}
Our starting point is SF~\citep{jiang2025streaming}. SF predicts the future by
starting from the most recent observation $\vx_m$ and learning how the system
state should move forward in physical time. The state produced during the ODE
rollout is therefore the forecast itself: as the integration time advances,
the predicted state advances along the future trajectory.

Let $\vh=\vx_{a:m}$ denote the observed history and let $s\in[0,1]$ denote
normalized physical forecast time. Conditioned on $\vh$, we write the future
trajectory as
$\mathbf{X}^{\mathrm{f}}
=(\mathbf{X}^{\mathrm{f}}_s)_{s\in[0,1]}$,
with $\mathbf{X}^{\mathrm{f}}_0=\vx_m$.
During training, we observe a realization
$\boldsymbol{\xi}:[0,1]\rightarrow\mathcal X$ of this future trajectory.
At any time $s$, the trajectory gives us its state
$\boldsymbol{\xi}(s)$ and physical velocity
$\dot{\boldsymbol{\xi}}(s)$. SF uses these quantities to construct a velocity
target for the observed trajectory and nearby states:
\begin{equation}
    \mathbf{g}_{\boldsymbol{\xi}}(\vx,s)
    =
    \dot{\boldsymbol{\xi}}(s)
    -
    \kappa
    \bigl(
        \vx-\boldsymbol{\xi}(s)
    \bigr),
    \qquad
    \kappa>0.
    \label{eq:continuous_sfp_field}
\end{equation}
The second term pulls nearby states back toward the trajectory, creating a
contracting tube around it. On the trajectory itself, this correction vanishes,
so the target reduces to $\dot{\boldsymbol{\xi}}(s)$.

This off-trajectory behavior is what gives SF its flow matching
interpretation. Rather than specifying only the velocity along the observed
trajectory, the target field also determines how nearby states move. To see
this, consider a small cloud of initial states around the beginning of the
trajectory,
$\mathbf{Y}_0\sim
\mathcal N(\boldsymbol{\xi}(0),\sigma_0^2\mathbf I_d)$,
and evolve these states according to
$\dot{\mathbf{Y}}_s=
\mathbf g_{\boldsymbol{\xi}}(\mathbf{Y}_s,s)$.
The center of the cloud follows the observed future
$\boldsymbol{\xi}(s)$, while deviations from that center are contracted by the
correction term.
This evolution has the closed-form solution
$
\mathbf{Y}_s
=
\boldsymbol{\xi}(s)
+
e^{-\kappa s}
\bigl(
\mathbf{Y}_0-\boldsymbol{\xi}(0)
\bigr)$.
The displacement from the observed trajectory therefore decays exponentially
at rate $\kappa$. Since the initial displacement is Gaussian, the state at
physical time $s$ remains Gaussian
\begin{equation}
\mathbf{Y}_s
\mid
\boldsymbol{\xi}
\sim
\mathcal N
\left(
\boldsymbol{\xi}(s),
\sigma_0^2 e^{-2\kappa s}\mathbf I_d
\right).
\label{eq:continuous_sfp_tube}
\end{equation}
Thus, each observed future trajectory defines a probability path around the
trajectory itself: a Gaussian tube whose mean follows
$\boldsymbol{\xi}(s)$ and whose width contracts over physical time.
The field $\mathbf g_{\boldsymbol{\xi}}$ is the velocity field associated with
this trajectory-conditioned path.

SF then learns a single velocity field
$\vv_{\vtheta}(\vx,s,\vh)$, represented by a neural network, from many such
future trajectories. Given the current state $\vx$, physical time $s$, and
observed history $\vh$, the network predicts how the state should move next.
Training matches this prediction to the velocity target associated with the
observed future. At the population level
\begin{equation}
    \mathcal L_{\mathrm{SF}}(\vtheta)
    =
    \mathbb E
    \left[
        \left\lVert
            \vv_{\vtheta}(\mathbf{Y}_s,s,\vh)
            -
            \mathbf{g}_{\mathbf{X}^{\mathrm{f}}}(\mathbf{Y}_s,s)
        \right\rVert_2^2
    \right],
    \label{eq:continuous_sf_objective}
\end{equation}
where $\mathbf{Y}_s$ denotes a state in the contracting tube around the
sampled future trajectory. In practice, training can be performed directly
on the observed trajectory. In that case,
$\mathbf{Y}_s=\boldsymbol{\xi}(s)$ and the target reduces to its
physical-time velocity $\dot{\boldsymbol{\xi}}(s)$.

\textbf{Discrete observations.}
The above formulation is continuous in time. In practice, however, we only observe the system state at discrete time points. Consider a training window $\vx_{a:T}$ with forecast origin $m$, and let
$P=T-m$ be the number of future transitions. We normalize the observation
times as $s_k=k/P$.
Between two consecutive observations, we use the piecewise-linear path
\begin{equation}
    \boldsymbol{\xi}^{\Delta}(s)
    =
    (1-\alpha_s)\vx_{m+k}
    +
    \alpha_s\vx_{m+k+1},
    \qquad
    \alpha_s=P(s-s_k),
    \label{eq:physical_time_interpolation}
\end{equation}
for $s\in[s_k,s_{k+1})$. Its physical-time velocity is constant within each
interval,
\begin{equation}
    \dot{\boldsymbol{\xi}}^{\Delta}(s)
    =
    P
    \bigl(
        \vx_{m+k+1}-\vx_{m+k}
    \bigr)
    =:
    \vu_k.
    \label{eq:physical_time_velocity}
\end{equation}
Thus, a discrete transition provides both a local state along the interpolated
trajectory and its physical-time velocity. We use these quantities to train
VSF in \cref{sec:method}.


\section{Variational Streaming Flow}
\label{sec:method}

\begin{figure}[!t]
  \centering
  \includegraphics[
    width=\linewidth,
    trim={5bp 0bp 0bp 5bp},
    clip
  ]{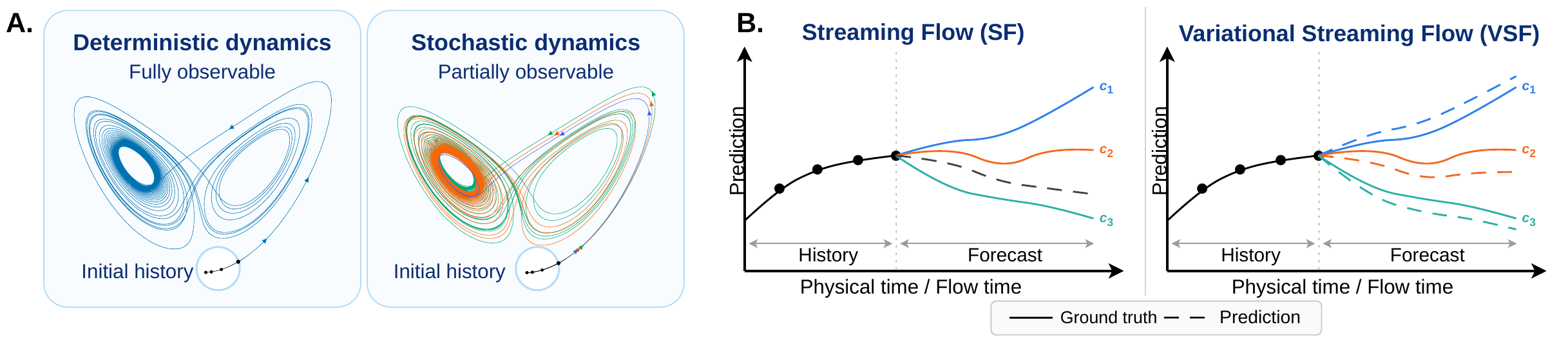}
  \vspace{-0.7cm}
  \caption{Deterministic and stochastic systems pose different forecasting
  problems. (A) When deterministic dynamics are fully observed, the future is
  unique. With partial observations and stochastic dynamics, the same observed
  history can lead to different future trajectories. (B) SF produces a single
  forecast from the observed state, while VSF can generate multiple plausible
  futures.}
  \label{fig:forecasting-settings}
  \vspace{-1em}
\end{figure}

Standard non-latent SF assigns a single physical-time velocity to a given
state and observed history, and therefore produces a single future from a fixed initial condition. This is sufficient when the future is unique, but not when the same observed history is compatible with different trajectories.
Figure~\ref{fig:forecasting-settings} illustrates this setting.
The difficulty becomes clear near a bifurcation. Different future
trajectories can  \emph{share the same physical trajectory} before separating, so SF~\citep{jiang2025streaming} must encode the information needed to select a future branch through small
differences in the physical state. As these differences shrink, the velocity
field must distinguish similar states, which can make the rollout more
sensitive to perturbations. App.~\ref{app_motivate} studies this
fidelity--sensitivity tradeoff in detail.

VSF addresses this limitation by allowing the physical-time velocity field to
depend on an additional latent variable. Let $\vh=\vx_{a:m}$ denote the
observed history and let $\vz\in\mathbb R^{d_z}$ denote the latent variable.
We condition the velocity field on $\vz$ and define the forecast dynamics as
\begin{tcolorbox}[
colback=parchment,
colframe=sandborder,
boxrule=0.3pt,
left=8pt,
right=8pt,
top=3pt,
bottom=3pt
]
\begin{equation}
\frac{d\widehat{\vx}_s}{ds}
=
\vv_{\vtheta}
\left(
\widehat{\vx}_s,s,\vh,\vz
\right).
\label{eq:vsf_dynamics}
\end{equation}
\end{tcolorbox}
Here, $s\in[0,1]$ is normalized physical forecast time and $\widehat{\vx}_0=\vx_m$. For a fixed $\vz$, the velocity field defines one physical-time trajectory starting from the most recent observation. Changing $\vz$ changes the velocity field along the rollout and can therefore produce a different future from the same observed history. We call this model \emph{Variational Streaming Flow} (VSF).

\subsection{Training and Inference}
\label{sec:vsf-training-inference}

VSF uses the latent variable differently during training and inference. During
training, the future trajectory is observed, so we can use its local motion to
infer which latent values explain the dynamics. At inference, the
future is unknown, so the latent must instead be sampled before the rollout.
We use the fixed prior
\begin{equation}
p(\vz)
=
\mathcal N
\left(
\mathbf{0},
\mathbf{I}_{d_z}
\right).
\label{eq:vsf_prior}
\end{equation}

We first describe training. As in \cref{sec:preliminaries}, we sample a
physical time $s$ from an observed future trajectory and construct the local
state $\vx_s$ together with its physical-time velocity $\vu_s$. The pair
$(\vx_s,\vu_s)$ tells us both where the trajectory is and how it is moving at
that time. We use an encoder, parameterized by $\vphi$, to map this local
future information together with $s$ and the observed history $\vh$ to the
parameters of a Gaussian distribution over $\vz$. Specifically, the encoder
outputs a mean
$\boldsymbol{\mu}_{\vphi}(\vu_s,\vx_s,s,\vh)\in\mathbb R^{d_z}$
and a vector of positive standard deviations
$\boldsymbol{\sigma}_{\vphi}(\vu_s,\vx_s,s,\vh)\in\mathbb R_{>0}^{d_z}$.
These outputs define the variational posterior
\begin{equation}
q_{\vphi}
\bigl(
\vz
\mid
\vu_s,\vx_s,s,\vh
\bigr)
=
\mathcal N
\left(
\boldsymbol{\mu}_{\vphi}(\vu_s,\vx_s,s,\vh),
\operatorname{diag}
\left[
\boldsymbol{\sigma}_{\vphi}^2(\vu_s,\vx_s,s,\vh)
\right]
\right).
\label{eq:vsf_variational_posterior}
\end{equation}

We draw $M$ latent samples from this posterior. Each sample is passed to the
velocity field together with the same local state, time, and history,
producing
$\widehat{\vu}_s^{(i)}
=
\vv_{\vtheta}(\vx_s,s,\vh,\vz^{(i)})$.
These predictions represent the physical-time velocities associated with
different latent values at the observed transition. Training encourages this
set of predicted velocities to match the observed local dynamics through the
distributional reconstruction loss $\mathcal L_{\mathrm{ES}}$. A KL term
simultaneously keeps the inferred posterior close to the prior in
\cref{eq:vsf_prior}, so that samples from this prior can be used when the
future is no longer available. We derive both terms in the following
subsections.

At inference, $\vu_s$ is unknown. Thus, we sample one latent $\vz\sim p(\vz)$ at the beginning of the forecast and use it throughout the rollout. Starting from
$\widehat{\vx}_0=\vx_m$, we discretize the physical-time dynamics in
\cref{eq:vsf_dynamics}. With a forward-Euler step of size $\Delta t$,
the update is
\begin{equation}
\widehat{\vx}_{k+1}
=
\widehat{\vx}_{k}
+
\Delta t\,
\vv_{\vtheta}
\left(
\widehat{\vx}_{k},
s_k,
\vh,
\vz
\right),
\label{eq:vsf_inference_update}
\end{equation}
where $s_k$ is the normalized physical time at the current forecast step.
The generated state $\widehat{\vx}_{k+1}$ is then added to the rolling
history used by the predictor. When the history representation $\vh$ is
refreshed from these generated states, the latent $\vz$ remains unchanged.
Thus, the history evolves with the generated trajectory while one latent draw
continues to condition the complete forecast.

The two procedures are summarized below. The complete algorithms, including
batching, history-window updates, and segment boundaries, are given in
App.~\ref{sec:vsf-forecasting}.

\noindent
\begin{minipage}[t]{0.485\linewidth}
\vspace{0pt}
\begin{tcolorbox}[
colback=sage,
colframe=sageborder,
boxrule=0.3pt,
left=6pt,
right=6pt,
top=5pt,
bottom=5pt,
title=\textbf{Training},
equal height group=vsf-summary
]
\small

Construct a local state and velocity
$(\vx_s,\vu_s)$.
\[
\vz^{(1:M)}
\stackrel{\mathrm{iid}}{\sim}
q_{\vphi}
\bigl(
\,\cdot\mid
\vu_s,\vx_s,s,\vh
\bigr)
\]
Evaluate the physical-time velocities
\[
\widehat{\vu}_s^{(1:M)}
=
\vv_{\vtheta}
\bigl(
\vx_s,s,\vh,\vz^{(1:M)}
\bigr)
\]
Train $\vv_{\vtheta}$ and the encoder using
$\mathcal L_{\mathrm{ES}}$ together with KL regularization toward $p(\vz)$.

\end{tcolorbox}
\end{minipage}
\hfill
\begin{minipage}[t]{0.485\linewidth}
\vspace{0pt}
\begin{tcolorbox}[
colback=lavender,
colframe=lavenderborder,
boxrule=0.3pt,
left=6pt,
right=6pt,
top=5pt,
bottom=5pt,
title=\textbf{Inference},
equal height group=vsf-summary
]
\small

Sample one latent from the prior
\[
\vz\sim p(\vz)
\]
and keep it fixed for the complete forecast.

Advance the state in physical time
\[
\widehat{\vx}_{k+1}
=
\widehat{\vx}_{k}
+
\Delta t\,
\vv_{\vtheta}
\bigl(
\widehat{\vx}_{k},s_k,\vh,\vz
\bigr)
\]
Add generated states to the rolling history and refresh $\vh$ while retaining
the same $\vz$.

\end{tcolorbox}
\end{minipage}

\subsection{Variational Training Objective}
\label{sec:vsf-objective}

We now derive a variational objective for the training procedure above. At a
sampled physical time $s$, the model observes the local state $\vx_s$, the
forecast time $s$, and the history $\vh$, and predicts the corresponding
physical-time velocity $\vu_s$. To keep the notation compact, define the local
context
\begin{equation}
\vc_s := (\vx_s,s,\vh).
\label{eq:vsf_local_context}
\end{equation}
The latent variable $\vz$ provides additional information about which local
dynamics are being realized.

We begin with an auxiliary probabilistic model for the local velocity. Given
the context $\vc_s$ and latent $\vz$, we place an isotropic Gaussian around
the velocity predicted by the VSF field
\begin{equation}
p_{\vtheta}
\bigl(
\vu_s
\mid
\vc_s,\vz
\bigr)
=
\mathcal N
\left(
\vu_s;
\vv_{\vtheta}
\bigl(
\vx_s,s,\vh,\vz
\bigr),
\sigma_u^2\mathbf I_d
\right),
\qquad
\sigma_u>0.
\label{eq:gaussian_velocity_likelihood}
\end{equation}
Here, $\sigma_u^2$ controls the variance of the Gaussian likelihood and
$\mathbf I_d$ is the identity matrix in the velocity dimension. This
likelihood is used only to formulate learning at a sampled physical time. It
models the local velocity $\vu_s$ and should not be interpreted as a Gaussian
model for the full forecast trajectory generated by \cref{eq:vsf_dynamics}.
Because $\vz$ is latent, the conditional velocity distribution is obtained by
integrating over the prior
\begin{equation}
p_{\vtheta}
\bigl(
\vu_s
\mid
\vc_s
\bigr)
=
\int
p_{\vtheta}
\bigl(
\vu_s
\mid
\vc_s,\vz
\bigr)
p(\vz)
\,d\vz.
\label{eq:vsf_velocity_evidence}
\end{equation}
Directly optimizing this marginal likelihood requires integrating over the
latent variable. During training, we instead use the encoder introduced in \cref{sec:vsf-training-inference} to approximate the posterior over
$\vz$ from the observed local dynamics. Applying the standard variational
bound gives
\begin{align*}
\log
p_{\vtheta}
\bigl(
\vu_s
\mid
\vc_s
\bigr)
\geq{}
\mathbb E_{
\mathbf Z\sim
q_{\vphi}(\cdot\mid\vu_s,\vc_s)
}
\left[
\log
p_{\vtheta}
\bigl(
\vu_s
\mid
\vc_s,\mathbf Z
\bigr)
\right]
\notag -
\operatorname{KL}
\left(
q_{\vphi}
\bigl(
\vz
\mid
\vu_s,\vc_s
\bigr)
\,\Vert\,
p(\vz)
\right).
\end{align*}

The two terms have the same roles as in the training procedure described
above. The first rewards latent samples for which the velocity field explains
the observed physical-time velocity. The second keeps the encoder distribution
close to the prior that will be used once the future velocity is unavailable
at inference.
For the Gaussian likelihood in \cref{eq:gaussian_velocity_likelihood}, the negative expected log-likelihood has a simple form. Up to an additive constant
\begin{align*}
-
\mathbb E_{
\mathbf Z\sim
q_{\vphi}(\cdot\mid\vu_s,\vc_s)
}
\left[
\log
p_{\vtheta}
\bigl(
\vu_s
\mid
\vc_s,\mathbf Z
\bigr)
\right]
\notag =
\frac{1}{2\sigma_u^2}
\mathbb E_{
\mathbf Z\sim
q_{\vphi}(\cdot\mid\vu_s,\vc_s)
}
\left[
\left\|
\vv_{\vtheta}
\bigl(
\vx_s,s,\vh,\mathbf Z
\bigr)
-
\vu_s
\right\|_2^2
\right]
+
\mathrm{const}.
\end{align*}
Thus, the local ELBO leads to a latent-conditioned MSE reconstruction together
with KL regularization toward the prior. This provides a natural variational
starting point for VSF.

\subsection{Distributional Velocity Reconstruction}
\label{sec:distributional-reconstruction}

The variational objective above leads naturally to an MSE reconstruction of
the local physical-time velocity. This is appropriate when the latent variable
fully separates the different future dynamics. In practice, however, different
future continuations can still remain compatible with the same context and
latent value. To make this precise, let $(\mathbf C,\mathbf U)$ denote the random local context and velocity obtained by sampling a training trajectory and a physical time, and let
\[
\mathbf Z
\sim
q_{\vphi}
\bigl(
\,\cdot\mid
\mathbf U,\mathbf C
\bigr).
\]
For a fixed encoder $q_{\vphi}$, the population minimizer of the MSE
reconstruction is
\begin{equation}
\vv^*(\vc,\vz)
=
\mathbb E
\left[
\mathbf U
\,\middle|\,
\mathbf C=\vc,\mathbf Z=\vz
\right].
\label{eq:generic_mse_conditional_mean}
\end{equation}
Thus, the learned velocity is the conditional mean of all local velocities that
remain compatible with the same $(\vc,\vz)$. If the encoder assigns distinct
future continuations to well-separated latent values, this conditional
distribution can be narrow. If different continuations still overlap in latent
space, their remaining velocity variation is averaged. Such overlap can arise
from imperfect inference, overlapping posterior distributions, or posterior
collapse.

We therefore train the velocity field using several latent samples jointly.
For the observed local transition $(\vx_s,\vu_s)$, draw
$\vz^{(1)},\ldots,\vz^{(M)}$ independently from the encoder and evaluate
\begin{equation}
\widehat{\vu}_s^{(i)}
=
\vv_{\vtheta}
\left(
\vx_s,s,\vh,\vz^{(i)}
\right),
\qquad
\vz^{(i)}
\stackrel{\mathrm{iid}}{\sim}
q_{\vphi}
\bigl(
\,\cdot\mid
\vu_s,\vx_s,s,\vh
\bigr).
\label{eq:posterior_velocity_samples}
\end{equation}
These samples give a set of latent-conditioned velocity predictions for the
same observed transition. Rather than matching each prediction independently
with MSE, we compare this set with the observed velocity using a generalized
energy reconstruction,
\begin{equation}
\mathcal L_{\mathrm{ES}}(s)
=
\frac{1}{M}
\sum_{i=1}^{M}
\left\|
\widehat{\vu}_s^{(i)}-\vu_s
\right\|_2^\alpha
-
\frac{1}{2M^2}
\sum_{i,j=1}^{M}
\left\|
\widehat{\vu}_s^{(i)}
-
\widehat{\vu}_s^{(j)}
\right\|_2^\alpha,
\qquad
0<\alpha<2.
\label{eq:vsf_energy_score}
\end{equation}
The first term keeps the predicted velocities close to the observed physical
velocity. The second term accounts for the spread among the predictions, so
the loss does not favor collapsing all latent samples to the same velocity.
For $\alpha=1$, \cref{eq:vsf_energy_score} has the standard energy-score
form and reduces to the CRPS in one dimension~\citep{gneiting2007strictly, lang2026aifs}.

This reconstruction also connects the posterior samples used during training
to the prior samples used at inference. Under aggregate posterior--prior
alignment, App.~\ref{sec:population_energy_control} shows that the expected
population energy reconstruction upper-bounds one half of the energy distance
between the prior-predictive velocity distribution and the true conditional
velocity distribution.
Combining the energy reconstruction with the KL regularization gives the objective that we use to train VSF
\begin{equation}
\mathcal L_{\mathrm{VSF}}
=
\mathbb E
\left[
\mathcal L_{\mathrm{ES}}(s)
+
\beta_{\mathrm{KL}}
\operatorname{KL}
\left(
q_{\vphi}
(\vz\mid\vu_s,\vx_s,s,\vh)
\,\Vert\,
p(\vz)
\right)
\right].
\label{eq:vsf_training_objective}
\end{equation}
The expectation is over training trajectories and sampled physical times.
Replacing the Gaussian ELBO reconstruction with
\cref{eq:vsf_energy_score} means that
\cref{eq:vsf_training_objective} is no longer an ELBO, but a KL-regularized
distributional reconstruction under the same local latent model.

\section{Experimental Results}
\label{sec:results}

In this section, we evaluate VSF on fully observed deterministic systems
(\Cref{sec:exp-deterministic}), partially observed stochastic systems
(\Cref{sec:exp-stochatic}), and as a predictor for model-based planning
(\Cref{sec:exp-world-model}). Additional experimental details and ablations are
provided in App.~\ref{sec:metrics}--\ref{sec:world-model-latent-trajectory-visualizations}.

\textbf{Baselines.} For the dynamical-system forecasting experiments, we compare VSF with eight baselines: (1) Block diffusion, a conditional diffusion model that generates the forecast horizon jointly; (2) AR diffusion, an autoregressive diffusion model that generates one future state at a time; (3) Rectified flow, which autoregressively generates one future state at a time by integrating a rectified-flow velocity field~\citep{liu2022flow}; (4) DeepAR~\citep{salinas2020deepar}, an autoregressive recurrent probabilistic
forecaster; (5) SF~\citep{jiang2025streaming}, a deterministic
physical-time flow; (6) SF (perturbation), which introduces
forecast diversity through Gaussian perturbations of the initial state; (7)
Chronos-2~\citep{ansari2025chronos}, a pretrained universal time-series
forecaster; and (8) Chronos-2 (calibrated), which applies post-hoc calibration
to the Chronos-2 predictions.

\begin{table}[t]
\centering
\vspace{-2em}
\caption{VSF achieves the lowest forecasting error across all deterministic
and stochastic systems over a 1,000-step rollout. We report CRPS, RMSE, and
MAE for deterministic systems and CME, CSE, $W_1$, and ED for stochastic
systems. Runtime (RT), in seconds, is averaged across tasks.}
\label{tab:forecasting_results}
\begingroup
\normalsize
\setlength{\tabcolsep}{2pt}

\resizebox{\linewidth}{!}{%
\begin{tabular}{@{}l*{3}{>{\columncolor{resultDPBg}}c}*{3}{>{\columncolor{resultRosslerBg}}c}*{3}{>{\columncolor{resultLorenzBg}}c}>{}r@{}}
\toprule
\multicolumn{11}{c}{\cellcolor{gray!10}{Deterministic dynamics}}\\
& \multicolumn{3}{c}{\cellcolor{resultDPBg}{Double pendulum}}
& \multicolumn{3}{c}{\cellcolor{resultRosslerBg}{R\"ossler}}
& \multicolumn{3}{c}{\cellcolor{resultLorenzBg}{Lorenz}}
& \\
\cmidrule(lr){2-4}\cmidrule(lr){5-7}\cmidrule(lr){8-10}
Model
& CRPS $\downarrow$ & RMSE $\downarrow$ & MAE $\downarrow$
& CRPS $\downarrow$ & RMSE $\downarrow$ & MAE $\downarrow$
& CRPS $\downarrow$ & RMSE $\downarrow$ & MAE $\downarrow$
& RT \\
\midrule
Block diffusion
& \resultpm{0.46}{0.07} & \resultpm{0.91}{0.14} & \resultpm{0.47}{0.07}
& \resultpm{0.29}{0.13} & \resultpm{0.54}{0.26} & \resultpm{0.29}{0.13}
& \resultpm{0.31}{0.12} & \resultpm{0.42}{0.18} & \resultpm{0.33}{0.12}
& 30.2 \\

AR diffusion
& \resultpm{0.43}{0.00} & \resultpm{0.96}{0.01} & \resultpm{0.59}{0.01}
& \resultpm{0.35}{0.06} & \resultpm{1.00}{0.13} & \resultpm{0.53}{0.08}
& \resultpm{0.86}{0.04} & \resultpm{1.96}{0.06} & \resultpm{1.50}{0.06}
& 78.2 \\

Rectified flow
& \resultpm{0.53}{0.07} & \resultpm{1.08}{0.11} & \resultpm{0.67}{0.07}
& \resultpm{0.40}{0.07} & \resultpm{0.85}{0.20} & \resultpm{0.48}{0.08}
& \resultpm{1.46}{0.89} & \resultpm{2.69}{1.46} & \resultpm{1.94}{1.14}
& 41.1 \\

DeepAR
& \resultpm{0.58}{0.00} & \resultpm{1.07}{0.00} & \resultpm{0.68}{0.00}
& \resultpm{0.17}{0.04} & \resultpm{0.54}{0.05} & \resultpm{0.21}{0.05}
& \resultpm{10.30}{0.22} & \resultpm{12.42}{0.13} & \resultpm{10.88}{0.13}
& 3.9 \\

SF
& \resultpm{2.12}{0.87} & \resultpm{9.18}{5.17} & \resultpm{2.12}{0.87}
& \resultpm{0.16}{0.07} & \resultpm{0.34}{0.18} & \resultpm{0.16}{0.07}
& \resultpm{1.42}{0.14} & \resultpm{2.75}{0.31} & \resultpm{1.42}{0.14}
& 6.4 \\

SF (perturb.)
& \resultpm{1.83}{0.86} & \resultpm{8.38}{5.18} & \resultpm{1.96}{0.89}
& \resultpm{0.11}{0.05} & \resultpm{0.34}{0.18} & \resultpm{0.16}{0.07}
& \resultpm{1.39}{0.14} & \resultpm{2.75}{0.31} & \resultpm{1.42}{0.14}
& 6.4 \\

Chronos-2
& \resultpm{0.78}{0.00} & \resultpm{1.50}{0.02} & \resultpm{0.96}{0.01}
& \resultpm{1.98}{0.00} & \resultpm{4.11}{0.02} & \resultpm{2.62}{0.01}
& \resultpm{13.50}{0.06} & \resultpm{28.19}{1.14} & \resultpm{25.22}{0.85}
& 0.8 \\

Chronos-2 (calib.)
& \resultpm{0.62}{0.00} & \resultpm{1.16}{0.00} & \resultpm{0.74}{0.00}
& \resultpm{1.91}{0.00} & \resultpm{3.91}{0.02} & \resultpm{2.52}{0.01}
& \resultpm{2.39}{0.02} & \resultpm{6.01}{0.27} & \resultpm{4.54}{0.27}
& 0.4 \\

VSF (Ours)
& \fullbestresultpm{0.25}{0.07} & \fullbestresultpm{0.71}{0.11} & \fullbestresultpm{0.33}{0.08}
& \fullbestresultpm{0.03}{0.00} & \fullbestresultpm{0.13}{0.01} & \fullbestresultpm{0.05}{0.00}
& \fullbestresultpm{0.16}{0.02} & \fullbestresultpm{0.25}{0.02} & \fullbestresultpm{0.19}{0.01}
& 7.2 \\
\bottomrule
\end{tabular}%
}

\vspace{-1.3pt}

\resizebox{\linewidth}{!}{%
\begin{tabular}{@{}l*{4}{>{\columncolor{resultRosslerBg}}c}*{4}{>{\columncolor{resultLorenzBg}}c}>{}r@{}}
\toprule
\multicolumn{10}{c}{\cellcolor{gray!10}{Stochastic dynamics}}\\
& \multicolumn{4}{c}{\cellcolor{resultRosslerBg}{R\"ossler}}
& \multicolumn{4}{c}{\cellcolor{resultLorenzBg}{Lorenz}}
& \\
\cmidrule(lr){2-5}\cmidrule(lr){6-9}
Model
& CME $\downarrow$ & CSE $\downarrow$ & $W_1$ $\downarrow$ & ED $\downarrow$
& CME $\downarrow$ & CSE $\downarrow$ & $W_1$ $\downarrow$ & ED $\downarrow$
& RT  \\
\midrule
Block diffusion
& \resultpm{0.16}{0.02} & \resultpm{0.72}{0.08} & \resultpm{8.88}{1.05} & \resultpm{12.67}{1.71}
& \resultpm{0.20}{0.01} & \resultpm{0.93}{0.01} & \resultpm{19.88}{0.28} & \resultpm{13.73}{0.45}
& 21.8 \\

AR diffusion
& \resultpm{0.20}{0.03} & \resultpm{4.63}{0.40} & \resultpm{16.21}{1.12} & \resultpm{11.46}{2.06}
& \resultpm{0.22}{0.05} & \resultpm{0.95}{0.14} & \resultpm{25.64}{3.10} & \resultpm{5.60}{1.54}
& 42.4 \\

Rectified flow
& \resultpm{0.22}{0.09} & \resultpm{0.89}{0.30} & \resultpm{12.60}{4.44} & \resultpm{16.72}{8.72}
& \resultpm{0.26}{0.03} & \resultpm{0.60}{0.12} & \resultpm{19.74}{1.63} & \resultpm{13.24}{2.69}
& 19.4 \\

DeepAR
& \resultpm{0.20}{0.02} & \resultpm{0.61}{0.02} & \resultpm{11.81}{0.80} & \resultpm{15.62}{1.57}
& \resultpm{2.14}{0.49} & \resultpm{2.61}{0.27} & \resultpm{131.84}{15.99} & \resultpm{138.30}{45.71}
& 1.2 \\

SF
& \resultpm{0.06}{0.02} & \resultpm{1.00}{0.00} & \resultpm{4.53}{0.67} & \resultpm{5.00}{1.34}
& \resultpm{1.04}{0.56} & \resultpm{1.00}{0.00} & \resultpm{58.12}{24.23} & \resultpm{89.84}{48.45}
& 7.5 \\

SF (perturb.)
& \resultpm{0.06}{0.02} & \resultpm{1.12}{0.23} & \resultpm{4.92}{0.84} & \resultpm{2.28}{0.82}
& \resultpm{0.68}{0.33} & \resultpm{1.33}{0.75} & \resultpm{53.62}{25.69} & \resultpm{35.78}{19.55}
& 6.6 \\

Chronos-2
& \resultpm{1.28}{0.06} & \resultpm{47.87}{1.63} & \resultpm{90.54}{3.86} & \resultpm{94.84}{0.79}
& \resultpm{6.84}{0.17} & \resultpm{37.49}{1.66} & \resultpm{411.32}{9.43} & \resultpm{342.98}{1.87}
& 0.3 \\

Chronos-2 (calib.)
& \resultpm{1.20}{0.06} & \resultpm{61.83}{2.09} & \resultpm{101.69}{4.61} & \resultpm{93.23}{0.20}
& \resultpm{1.44}{0.06} & \resultpm{10.53}{0.49} & \resultpm{111.81}{3.11} & \resultpm{64.97}{0.79}
& 0.1 \\

VSF (Ours)
& \fullbestresultpm{0.04}{0.01} & \fullbestresultpm{0.36}{0.14} & \fullbestresultpm{2.62}{0.38} & \fullbestresultpm{1.28}{0.42}
& \fullbestresultpm{0.19}{0.03} & \fullbestresultpm{0.50}{0.21} & \fullbestresultpm{15.95}{5.64} & \fullbestresultpm{5.19}{0.93}
& 6.7 \\
\bottomrule
\end{tabular}%
}

\endgroup
\end{table}

\vspace{-1em}
\begin{figure}[t]
  \centering
\includegraphics[trim={0bp 0bp 0bp 2bp}, clip, width=0.88\linewidth]{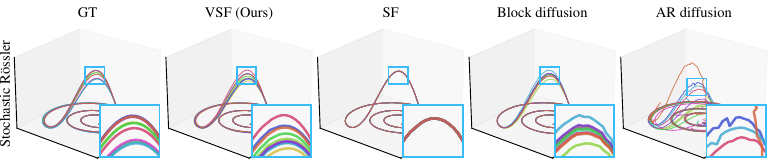}
\caption{VSF preserves the R\"ossler geometry while capturing variability
across plausible futures. We show $10$ trajectories from each method. Block
diffusion produces less diverse forecasts, while AR diffusion is more prone to
rollout drift.}
  \label{fig:qualitative-stochastic-rossler}
  \label{fig:qualitative-stochastic}
  \vspace{-0.5cm}
\end{figure}

\subsection{Deterministic Forecasting} \label{sec:exp-deterministic}

We evaluate fully observed deterministic forecasting on the double pendulum,
R\"ossler, and Lorenz systems using 1,000-step rollouts. We report CRPS for the
forecast ensemble and RMSE and MAE for its mean; detailed definitions are
provided in App.~\ref{sec:metrics}.

\textbf{VSF produces accurate forecasts with samples concentrated around the
ground-truth trajectory.}
As shown in Table~\ref{tab:forecasting_results}, VSF achieves the lowest CRPS,
RMSE, and MAE across all three systems, while providing up to a $7\times$
speedup over AR diffusion.
The qualitative forecasts in the Appendix
(\cref{fig:main-chaotic-double-pendulum}) further show that its sampled
trajectories remain close to the reference trajectory throughout the rollout.
Chronos-2 is less accurate in these experiments. One possible
factor is the short forecasting context of only 10 time steps.
This is consistent with~\citet{zhang2025zero}, who use 512 context points in
their experiments on chaotic systems and report improved zero-shot
forecasting performance for Chronos as the context length increases.

\begin{wrapfigure}[12]{r}{0.51\linewidth}
  \vspace{-1.5em}
  \centering
  \begingroup
  \setlength{\tabcolsep}{1pt}
  \renewcommand{\arraystretch}{1.0}
  \newcommand{\stochasticpreview}[1]{%
    \includegraphics[width=0.3\linewidth]{#1}%
  }
  \newcommand{\stochasticpreviewheader}[1]{%
    {\normalfont\scriptsize\shortstack{#1}}%
  }
  \begin{tabular}{@{}*{3}{c}@{}}
    \stochasticpreviewheader{GT}
    & \stochasticpreviewheader{VSF}
    & \stochasticpreviewheader{AR diff.} \\[0.1em]
    \stochasticpreview{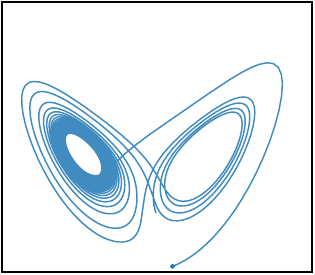}
    & \stochasticpreview{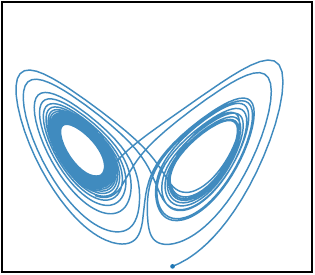}
    & \stochasticpreview{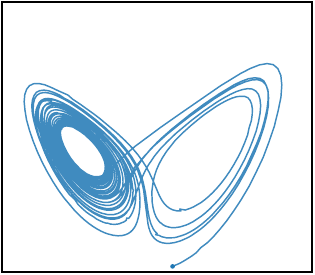} \\[0.1em]
    \stochasticpreview{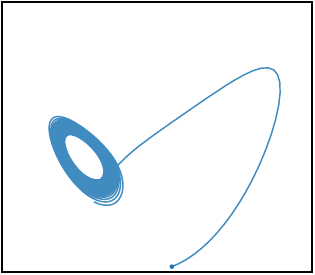}
    & \stochasticpreview{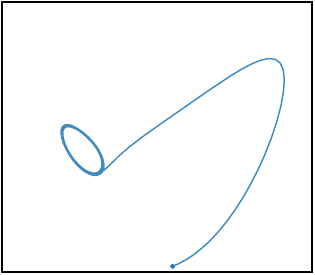}
    & \stochasticpreview{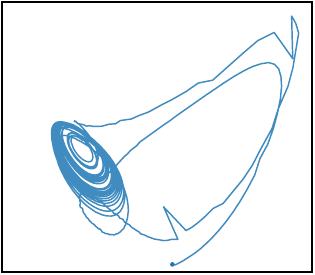}
  \end{tabular}
  \endgroup
    \vspace{-0.2cm}
    \caption{VSF samples remain consistent with the stochastic Lorenz dynamics,
while AR diffusion drifts from the attractor. Two futures are shown.}
  \label{fig:main-stochastic-lorenz-samples}
  \vspace{-0.8em}
\end{wrapfigure}
\textbf{VSF reduces long-horizon drift relative to SF.}
This difference is most apparent on the double pendulum in
\cref{fig:main-rollout-error-curves}. Adding noise to the initial state
of SF reduces the error somewhat, but its forecasts still drift progressively
from the reference trajectory. VSF maintains lower error over the
same rollout. In our implementation, the previous-frame representation is
combined with the latent variable $\vz$ through a weighted sum at the network
bottleneck (see \cref{sec:experimental-setup}). The ablation in
\cref{sec:full-ablation-results} shows that this latent conditioning is important
for the improved long-horizon accuracy.
\vspace{-3mm}

\subsection{Stochastic Forecasting} \label{sec:exp-stochatic}

We evaluate partially observed stochastic forecasting on the R\"ossler and
Lorenz systems. For each initial state, we generate multiple trajectories by
independently sampling an unobserved dynamical parameter. Each trajectory is
deterministic once this parameter is fixed, but the model does not observe it.
The same observed history can therefore lead to different future trajectories,
including trajectories that separate into different dynamical regimes.
We evaluate these forecast distributions using conditional mean error (CME),
conditional standard-deviation error (CSE), Wasserstein distance ($W_1$), and
energy distance (ED). For the values reported in our tables, $W_1$ and ED treat
each complete rollout as a single vector, so they compare distributions over
whole trajectories and account for temporal structure. In the rollout-error
plots, $W_1$ and ED are instead computed pointwise in time. Detailed metric
definitions are provided in App.~\ref{sec:metrics}.

\textbf{VSF best matches the conditional distribution of future trajectories.}
As shown in Table~\ref{tab:forecasting_results}, VSF achieves the lowest value
for every metric on both stochastic systems. The trajectory ensembles in
Figures~\ref{fig:qualitative-stochastic} and
\ref{fig:qualitative-stochastic-lorenz} show that VSF captures both the spread
of the reference futures and the geometry of the underlying dynamics. Block
diffusion and DeepAR cover a narrower part of the distribution, while AR
diffusion produces diverse futures but is more prone to rollout drift.
Perturbing the initial state of SF introduces some diversity, but this
inference-time perturbation is not learned by its training objective.
Unperturbed SF produces only a single trajectory, yielding a CSE of $1$.

%
\begin{wrapfigure}[12]{r}{0.55\linewidth}
  \vspace{-0.8em}
  \centering
  \setlength{\abovecaptionskip}{3pt}
  \captionof{table}{VSF achieves the lowest CME, CSE, and ED on stochastic
  Lorenz over a 2,500-step rollout. AR diffusion produces diverse samples but
  can drift from the system dynamics, as shown in
  \cref{fig:main-stochastic-lorenz-samples}.}
  \label{tab:stochastic-lorenz-2500}

  \begingroup
  \setlength{\tabcolsep}{2pt}
  \resizebox{\linewidth}{!}{%
  \scriptsize
  \begin{tabular}{@{}l*{4}{>{\columncolor{resultLorenzBg}}c}@{}}
  \toprule
  \multicolumn{5}{c}{\cellcolor{gray!10}{Stochastic dynamics}}\\
  & \multicolumn{4}{c}{\cellcolor{resultLorenzBg}{Lorenz}} \\
  \cmidrule(lr){2-5}
  Model & CME $\downarrow$ & CSE $\downarrow$ & $W_1$ $\downarrow$
  & ED $\downarrow$ \\
  \midrule
  Block diffusion
  & \resultpm{0.79}{0.09} & \resultpm{0.66}{0.11}
  & \fullbestresultpm{76.48}{3.20} & \resultpm{34.79}{11.90} \\
  AR diffusion
  & \resultpm{0.62}{0.03} & \resultpm{0.53}{0.03}
  & \resultpm{96.60}{2.48} & \resultpm{12.43}{0.52} \\
  Rectified flow
  & \resultpm{0.77}{0.16} & \resultpm{0.59}{0.21}
  & \resultpm{90.41}{1.75} & \resultpm{39.60}{16.90} \\
  SF (perturb.)
  & \resultpm{1.13}{0.33} & \resultpm{0.70}{0.21}
  & \resultpm{111.95}{22.66} & \resultpm{39.81}{14.86} \\
  VSF
  & \fullbestresultpm{0.56}{0.07} & \fullbestresultpm{0.39}{0.12}
  & \resultpm{85.67}{7.29} & \fullbestresultpm{12.00}{2.67} \\
  \bottomrule
  \end{tabular}%
  }
  \endgroup
  \vspace{-0.8em}
\end{wrapfigure}
\textbf{VSF preserves trajectory structure over very long horizons.}
For the 2,500-step Lorenz rollout in Table~\ref{tab:stochastic-lorenz-2500}, VSF achieves the lowest CME, CSE, and ED.
AR diffusion also achieves a low ED, but its sampled trajectories drift
from the system dynamics, as shown in
\cref{fig:main-stochastic-lorenz-samples}. Block diffusion reaches a lower
$W_1$, but its higher CSE and ED are consistent with the narrower mode coverage
seen in \cref{fig:qualitative-stochastic-lorenz}. 
In contrast, VSF
samples traverse both lobes of the attractor while preserving its
characteristic geometry.
\par

\subsection{World Model Predictor}
\label{sec:exp-world-model}

\begin{figure}[t]
  \centering
  \includegraphics[width=\linewidth, trim={0bp 260bp 0bp 30bp}, clip]{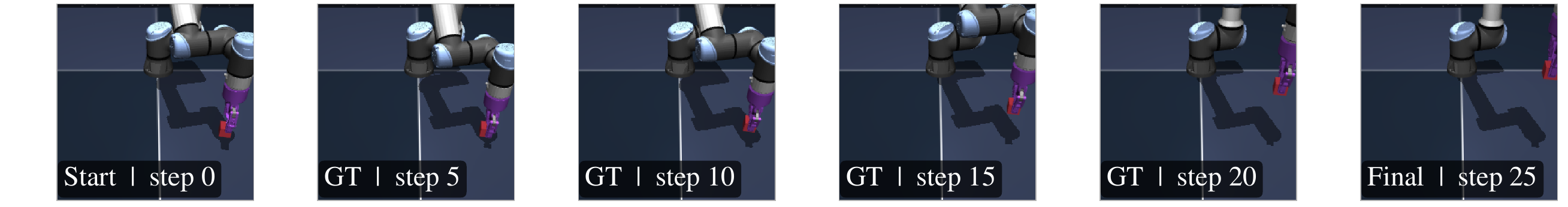}
  \includegraphics[width=\linewidth, trim={0bp 0bp 0bp 223bp}, clip]{figures/wm/cube_task_002_vsf_latent_trajectory.pdf}
  \vspace{-1.5em}
\caption{VSF serves as an effective world-model predictor on Cube. The top row
shows ground-truth rollout frames, and the bottom row compares the VSF latent
trajectory (red) with the ground truth (dashed). The prediction closely follows
the ground truth and leads to a successful plan.}
  \label{fig:wm-cube-task-035-vsf-latent-trajectory}
    \vspace{-0.2cm}
\end{figure}

We next evaluate VSF as an action-conditioned predictor in a JEPA-based world
model. The goal is to test whether its physical-time forecasting formulation
can improve prediction and planning in a learned representation space. We
freeze LeWM's pretrained JEPA representation~\citep{assran2023self, maes2026leworldmodel} and
train only the predictor on offline observation--action trajectories, following
the action-conditioned training setup of V-JEPA~2~\citep{assran2025v}. Future
latent states are used as teacher-forced training targets.
We compare LeWM's AR predictor with SF and VSF predictors while keeping the
JEPA encoder, CEM planner, and planning budget fixed. We report
latent-trajectory RMSE along the action sequences selected by CEM and planning
success on PushT, Cube, TwoRoom, and Reacher. The learned JEPA representation
also evolves smoothly over time, as illustrated in
\cref{fig:wm-cube-task-035-vsf-latent-trajectory}, making physical-time
prediction a natural fit for this latent space. Details of the
action-conditioned VSF extension are provided in
App.~\ref{sec:wm-vsf-algorithms}.

\textbf{VSF improves both latent prediction and planning success.}
As shown in \cref{tab:world-model-predictor-results}, VSF reduces
latent-trajectory RMSE relative to the AR predictor on all four tasks and
achieves the lowest RMSE on PushT, TwoRoom, and Reacher. It also achieves the
highest planning success on every task, improving average success by $7.34\%$
over the AR predictor. The successful Cube rollout in
\cref{fig:wm-cube-task-035-vsf-latent-trajectory} provides a qualitative
example, with the predicted latent trajectory closely following the ground
truth. VSF retains LeWM's budget of one velocity-field evaluation
per imagined transition.

\begin{table}[t]
\centering
\caption{As a world-model predictor, VSF achieves the highest planning success on all four tasks and the lowest predictor RMSE on three of four tasks. RMSE is measured between predicted and ground-truth latent trajectories along the CEM-selected action sequence.}
\label{tab:world-model-predictor-results}
\begingroup
\setlength{\tabcolsep}{4pt}
\resizebox{\linewidth}{!}{%
\begin{tabular}{@{}l*{8}{c}@{}}
\toprule
& \multicolumn{2}{c}{PushT}
& \multicolumn{2}{c}{Cube}
& \multicolumn{2}{c}{TwoRoom}
& \multicolumn{2}{c}{Reacher} \\
\cmidrule(lr){2-3}\cmidrule(lr){4-5}\cmidrule(lr){6-7}\cmidrule(lr){8-9}
Method
& Success & RMSE $\downarrow$
& Success & RMSE $\downarrow$
& Success & RMSE $\downarrow$
& Success & RMSE $\downarrow$ \\
\midrule
AR predictor
& \resultpm{84.7}{4.1} & \resultpm{0.28}{0.01}
& \resultpm{69.3}{2.5} & \resultpm{0.80}{0.05}
& \resultpm{78.7}{9.0} & \resultpm{0.67}{0.03}
& \resultpm{61.3}{8.2} & \resultpm{1.06}{0.01} \\

SF predictor
& \resultpm{88.0}{2.8} & \resultpm{0.30}{0.04}
& \resultpm{76.7}{0.9} & \fullbestresultpm{0.64}{0.02}
& \resultpm{76.7}{8.2} & \resultpm{0.59}{0.05}
& \resultpm{49.3}{5.0} & \resultpm{1.13}{0.03} \\

VSF predictor
& \fullbestresultpm{92.7}{2.5} & \fullbestresultpm{0.27}{0.03}
& \fullbestresultpm{80.0}{1.6} & \resultpm{0.65}{0.08}
& \fullbestresultpm{82.7}{5.0} & \fullbestresultpm{0.58}{0.01}
& \fullbestresultpm{68.0}{0.0} & \fullbestresultpm{1.03}{0.02} \\
\bottomrule
\end{tabular}%
}
\endgroup
\end{table}

\section{Conclusion}

We introduced VSF, a probabilistic extension of SF for forecasting directly in
physical time, where one latent variable conditions an entire rollout. This
allows VSF to model different possible futures while preserving the direct
physical-time generation of SF. We train the latent-conditioned velocity field
with variational inference and an energy-based reconstruction over multiple
latent samples. We also provide a population-level guarantee connecting this
objective to how well the model captures the distribution of possible local
dynamics under prior sampling. Empirically, VSF improves long-horizon
forecasting on deterministic and partially observed stochastic systems,
including the quality of the predicted future distribution, and improves
planning success as an action-conditioned world-model predictor, while
retaining one velocity-field evaluation per imagined transition.

\paragraph{Limitations and future work.} VSF relies on adjacent states providing a meaningful local direction of motion for learning the physical-time velocity. This assumption is most reliable when the dynamics vary smoothly in physical time. Abrupt impacts, discontinuities, or strong observation noise can make the local velocity poorly resolved. VSF also uses explicit forward Euler integration, which can accumulate discretization error for coarse step sizes or rapidly varying learned fields~\citep{hairer1993solving}; solver choice can also affect the effective dynamics learned by neural ODEs~\citep{zhu2022numerical}. Future work could therefore explore hybrid or jump-aware latent dynamics~\citep{jia2019neural},
controlled or rough differential equations~\citep{kidger2020neural,morrill2021neural}, and higher-order or
adaptive Runge--Kutta solvers~\citep{queiruga2020continuous}.

\subsection*{Acknowledgments}
HH, MG, NBE would like to acknowledge support from the U.S. Department of Energy, Office of Science, Office of Advanced Scientific Computing Research, through the EXPRESS: 2025 Exploratory Research for Extreme-Scale Science program and the RAPIDS Institute under the Scientific Discovery through Advanced Computing (SciDAC) program, under Contract No. DE-AC02-05CH11231 at Lawrence Berkeley National Laboratory. The views, opinions, and/or findings expressed are those of the authors and should not be interpreted as representing the official views or policies of the Department of Defense or the U.S. Government.
HH and PL would like to acknowledge the support by the National Science Foundation (NSF) under awards PHY-2117997,
IIS-2239565, IIS-2428777, and CCF-2402816; the Google Cloud Research Credit, 2026; the Nvidia Academic Award, 2026.
SHL would like to acknowledge support from the Wallenberg Initiative on Networks and Quantum Information (WINQ) and the Swedish Research Council (VR/2021-03648).

\bibliographystyle{iclr2027_conference}
\bibliography{iclr2027_conference}

\clearpage
\appendix

\section*{Appendix}

\section{Motivation for VSF}
\label{app_motivate}

A limitation of standard non-latent Streaming Flow (SF) \citep{jiang2025streaming} appears when multiple
future trajectories share (almost) the same physical state before
bifurcating. The following
symmetric two-branch example isolates this issue: even at the population optimum, standard SF can exhibit a fidelity-robustness tradeoff near a bifurcation.

\begin{figure}[!b]
    \centering
    \includegraphics[width=1.005\linewidth]{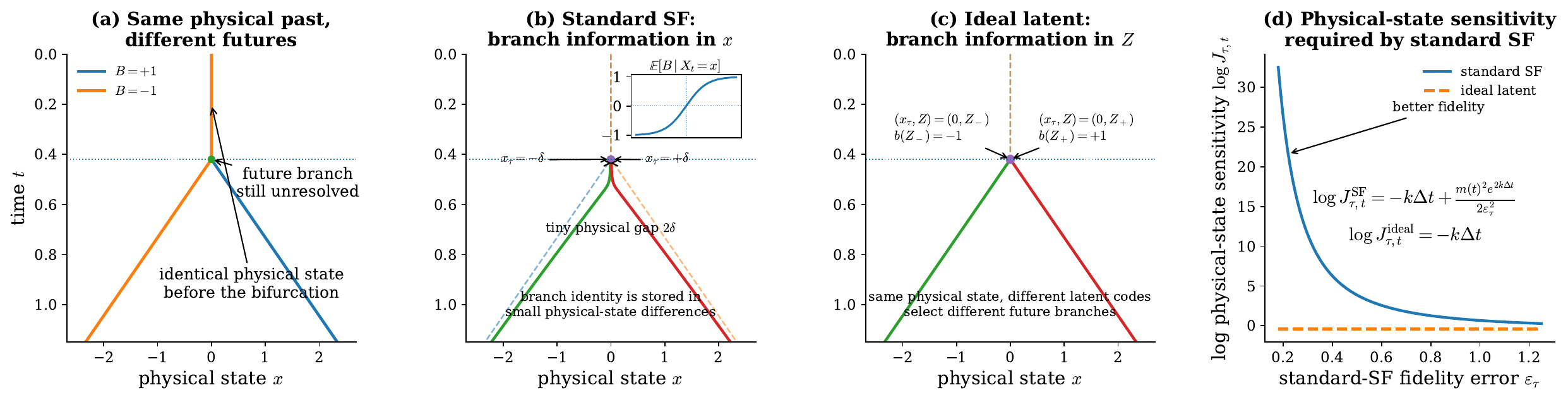}
            \caption{(a) Two futures share the same physical state up to the bifurcation time $\tau$.
(b) In non-latent SF, branch identity must be encoded by small differences in the physical state $x$, making the branch signal increasingly sharp as the Gaussian tube narrows.
(c) An ideal latent realization instead carries branch identity in $Z$, allowing the same physical state $X_\tau=0$ to generate either future.
(d) Accordingly, the standard-SF physical-state sensitivity grows as
$J_{\tau,t}^{\mathrm{SF}}\sim \exp(C/\varepsilon_\tau^2)$
as the  fidelity error $\varepsilon_\tau\to0$, while the ideal latent realization retains
$J_{\tau,t}^{\mathrm{ideal}}=e^{-k(t-\tau)}$. }
    \label{fig:motivate}
\end{figure}

{\bf A bifurcation example.} Fix an observed history $H=h$. Let $B\in\{-1,+1\}$ be equally likely
conditional on $h$, and consider the two future trajectories
\begin{equation}
    \xi_B(t)=B\,m(t),
    \qquad t\in[0,T],
    \label{eq:bifurcation_paths}
\end{equation}
where $m:[0,T]\to[0,\infty)$ is absolutely continuous on $[0,T]$,
continuously differentiable on $(\tau,T]$, satisfies
$m(t)=0$ for $t\leq\tau$, and $m(t)>0$ for $t>\tau$.
Thus the two futures share the same physical trajectory up to the
bifurcation time $\tau$ and separate afterward.

Let
$k>0$ and $\sigma_0>0$, and consider the branch-conditioned fields
\begin{equation}
    g_B(x,t)
    =
    \dot\xi_B(t)-k\bigl(x-\xi_B(t)\bigr).
    \label{eq:bifurcation_sf_conditional_field}
\end{equation}
Define
$\sigma_t^2=\sigma_0^2e^{-2kt},$
$q(t)=\dot m(t)+km(t)$.

\begin{vsfproposition}[Fidelity-sensitivity tradeoff]
\label{prop_motivate} Assume that the population optimum of the SF objective is attained.  Then, we have the following results.

(i) The optimal SF marginal field is
\begin{equation}
    v_{\mathrm{SF}}^\star(x,t)
    =
    -kx
    +
    q(t)
    \tanh\!\left(
        \frac{x\,m(t)}{\sigma_t^2}
    \right).
    \label{eq:bifurcation_sf_field}
\end{equation}
At the bifurcation time,
$P_{\tau}^{\mathrm{SF}}
    =
    \mathcal N(0,\sigma_\tau^2),$
    $P_{\tau}^{\mathrm{data}}
    =
    \delta_0,$
and therefore
\begin{equation}
    \varepsilon_\tau
    :=
    W_2\!\left(
        P_{\tau}^{\mathrm{SF}},
        P_{\tau}^{\mathrm{data}}
    \right)
    =
    \sigma_\tau.
    \label{eq:bifurcation_sf_fidelity}
\end{equation}

(ii) $x(t)\equiv0$ is a trajectory of
$v_{\mathrm{SF}}^\star$. 
  If
$\Phi_{\tau,t}^{\mathrm{SF}}$ denotes its flow map, then for every
$t>\tau$,
\begin{equation}
    J_{\tau,t}^{\mathrm{SF}}
    :=
    \left.
    \frac{\partial
        \Phi_{\tau,t}^{\mathrm{SF}}(x)}
        {\partial x}
    \right|_{x=0}
    =
    \exp\!\left[
        -k(t-\tau)
        +
        \frac{
            m(t)^2e^{2k(t-\tau)}
        }{
            2\varepsilon_\tau^2
        }
    \right].
    \label{eq:bifurcation_fidelity_sensitivity}
\end{equation}
Hence, for every fixed $t>\tau$ with $m(t)>0$, and with
$k$, $m$, and $\tau$ fixed,
\begin{equation}
    \sigma_0\downarrow0
    \quad\Longleftrightarrow\quad
    \varepsilon_\tau\downarrow0
    \quad\Longrightarrow\quad
    J_{\tau,t}^{\mathrm{SF}}\to\infty.
    \label{eq:bifurcation_tradeoff_limit}
\end{equation}
\end{vsfproposition} 
Thus, in this example, standard SF faces
an explicit fidelity-sensitivity tradeoff: making the physical tube
arbitrarily close to the common target state forces the population
marginal field near the branch separator to become arbitrarily sensitive
to physical-state perturbations.

\paragraph{Proof.}
For branch $B$, plugging $\xi_B(t)=Bm(t)$ into
\eqref{eq:bifurcation_sf_conditional_field} gives
\begin{equation}
    g_B(x,t)
    =
    -kx+Bq(t),
    \qquad
    q(t)=\dot m(t)+km(t).
    \label{eq:bifurcation_sf_branch_field}
\end{equation}

We first derive the branch-conditioned  marginal field. Let
$E_t=X_t-\xi_B(t)$. From
\eqref{eq:bifurcation_sf_conditional_field},
\begin{equation}
    \dot E_t
    =
    \dot X_t-\dot\xi_B(t)
    =
    -kE_t.
\end{equation}
Hence
\begin{equation}
    E_t=e^{-kt}E_0.
\end{equation}
Since $\xi_B(0)=0$ and the standard SF construction uses
$X_0\sim\mathcal N(0,\sigma_0^2)$, we have
$E_0\sim\mathcal N(0,\sigma_0^2)$. Consequently,
\begin{equation}
    X_t\mid B
    \sim
    \mathcal N\!\left(
        Bm(t),
        \sigma_t^2
    \right),
    \qquad
    \sigma_t^2=\sigma_0^2e^{-2kt}.
    \label{eq:bifurcation_sf_branch_marginal}
\end{equation}

Under conditional flow matching, the population-optimal
marginal field is the conditional expectation of the branch-conditioned
target:
\begin{equation}
    v_{\mathrm{SF}}^\star(x,t)
    =
    \mathbb E[
        g_B(X_t,t)\mid X_t=x],
    \label{eq:bifurcation_sf_conditional_mean}
\end{equation}
where the corresponding marginal probability path is:
\begin{equation}
    P_t^{\mathrm{SF}}
    =
    \frac12\mathcal N(m(t),\sigma_t^2)
    +
    \frac12\mathcal N(-m(t),\sigma_t^2).
    \label{eq:bifurcation_sf_mixture}
\end{equation}

Using \eqref{eq:bifurcation_sf_branch_field},
\begin{equation}
    v_{\mathrm{SF}}^\star(x,t)
    =
    -kx
    +
    q(t)\,
    \mathbb E[B\mid X_t=x].
    \label{eq:bifurcation_sf_branch_mean}
\end{equation}
It remains to evaluate the posterior mean of $B$. From
\eqref{eq:bifurcation_sf_branch_marginal} and the equal branch
probabilities,
\begin{equation}
    \frac{
        \mathbb P(B=+1\mid X_t=x)
    }{
        \mathbb P(B=-1\mid X_t=x)
    }
    =
    \frac{
        \exp\!\left(
            -(x-m(t))^2/(2\sigma_t^2)
        \right)
    }{
        \exp\!\left(
            -(x+m(t))^2/(2\sigma_t^2)
        \right)
    }.
\end{equation}
Therefore
\begin{align}
    \log
    \frac{
        \mathbb P(B=+1\mid X_t=x)
    }{
        \mathbb P(B=-1\mid X_t=x)
    }
    &=
    \frac{
        (x+m(t))^2-(x-m(t))^2
    }{
        2\sigma_t^2
    }
    \notag\\
    &=
    \frac{2x\,m(t)}{\sigma_t^2}.
\end{align}
For a random variable taking values in $\{-1,+1\}$, this gives:
\begin{equation}
    \mathbb E[B\mid X_t=x]
    =
    \tanh\!\left(
        \frac{x\,m(t)}{\sigma_t^2}
    \right).
\end{equation}
Substitution into
\eqref{eq:bifurcation_sf_branch_mean} allows us to obtain
\eqref{eq:bifurcation_sf_field}.

At the bifurcation time, $m(\tau)=0$. Hence
\eqref{eq:bifurcation_sf_branch_marginal} gives
\begin{equation}
    P_{\tau}^{\mathrm{SF}}
    =
    \mathcal N(0,\sigma_\tau^2),
\end{equation}
whereas the two noiseless target branches both satisfy
$\xi_B(\tau)=0$, so
\begin{equation}
    P_{\tau}^{\mathrm{data}}=\delta_0.
\end{equation}
The only coupling between a distribution and a Dirac mass sends every
sample to the location of the Dirac mass. Therefore
\begin{align}
    W_2^2\!\left(
        P_{\tau}^{\mathrm{SF}},
        \delta_0
    \right)
    &=
    \mathbb E|X_\tau|^2
    = \sigma_\tau^2.
\end{align}
Thus
$W_2\!\left(
        P_{\tau}^{\mathrm{SF}},
        P_{\tau}^{\mathrm{data}}
    \right)  =
    \sigma_\tau,$
which is precisely
\eqref{eq:bifurcation_sf_fidelity}.

We next compute the sensitivity of the population SF near
the branch separator. By symmetry,
$v_{\mathrm{SF}}^\star(0,t)=0,$
so $x(t)\equiv0$ is a trajectory. Differentiating
\eqref{eq:bifurcation_sf_field} with respect to $x$ gives
\begin{equation}
    \partial_xv_{\mathrm{SF}}^\star(x,t)
    =
    -k
    +
    \frac{
        m(t)q(t)
    }{
        \sigma_t^2
    }
    \operatorname{sech}^2\!\left(
        \frac{x\,m(t)}{\sigma_t^2}
    \right).
\end{equation}
Along the central trajectory,
\begin{equation}
    \partial_xv_{\mathrm{SF}}^\star(0,t)
    =
    -k
    +
    \frac{
        m(t)[\dot m(t)+km(t)]
    }{
        \sigma_t^2
    }.
    \label{eq:bifurcation_sf_central_jacobian}
\end{equation}

Let
\begin{equation}
    J_{\tau,t}^{\mathrm{SF}}
    =
    \left.
    \frac{\partial
        \Phi_{\tau,t}^{\mathrm{SF}}(x)}
        {\partial x}
    \right|_{x=0}.
\end{equation}
The variational equation for the one-dimensional ODE flow gives
\begin{equation}
    \frac{d}{dt}J_{\tau,t}^{\mathrm{SF}}
    =
    \partial_xv_{\mathrm{SF}}^\star(0,t)
    J_{\tau,t}^{\mathrm{SF}},
    \qquad
    J_{\tau,\tau}^{\mathrm{SF}}=1.
\end{equation}
Therefore
\begin{equation}
    J_{\tau,t}^{\mathrm{SF}}
    =
    \exp\!\left(
        \int_\tau^t
        \partial_xv_{\mathrm{SF}}^\star(0,s)\,ds
    \right).
    \label{eq:bifurcation_sf_variational_solution}
\end{equation}

Since $\sigma_s^2=\sigma_0^2e^{-2ks}$,
\begin{equation}
    \frac{d}{ds}
    \left(
        \frac{m(s)^2}{2\sigma_s^2}
    \right)
    =
    \frac{
        m(s)[\dot m(s)+km(s)]
    }{
        \sigma_s^2
    }.
\end{equation}
Using $m(\tau)=0$ in
\eqref{eq:bifurcation_sf_central_jacobian}, we obtain
\begin{align}
    \int_\tau^t
    \partial_xv_{\mathrm{SF}}^\star(0,s)\,ds
    &=
    -k(t-\tau)
    +
    \int_\tau^t
    \frac{
        m(s)[\dot m(s)+km(s)]
    }{
        \sigma_s^2
    }
    ds
    \notag\\
    &=
    -k(t-\tau)
    +
    \frac{m(t)^2}{2\sigma_t^2}.
\end{align}

From
$\sigma_t^2=\sigma_\tau^2e^{-2k(t-\tau)}$ and
$\varepsilon_\tau=\sigma_\tau$, we further obtain
\begin{equation}
    \frac{m(t)^2}{2\sigma_t^2}
    =
    \frac{
        m(t)^2e^{2k(t-\tau)}
    }{
        2\varepsilon_\tau^2
    }.
\end{equation}
Using these and
\eqref{eq:bifurcation_sf_variational_solution}, we  obtain
\eqref{eq:bifurcation_fidelity_sensitivity}. 

Finally, for every fixed
$t>\tau$ with $m(t)>0$, the second term in the exponent diverges as
$\varepsilon_\tau\to0$, which proves
\eqref{eq:bifurcation_tradeoff_limit}.

\begin{vsfcorollary}
\label{cor:sf_fidelity_robustness_lower_bound}
Under the assumptions of
Proposition~\ref{prop_motivate}, suppose that for
some fixed $t>\tau$ the SF sensitivity is required to
satisfy
$J_{\tau,t}^{\mathrm{SF}}\leq J_{\max}$ for some threshold $J_{\max}$. Then,
whenever this constraint is feasible, it is necessary that
$J_{\max}>e^{-k(t-\tau)}$, and
\begin{equation}
    \varepsilon_\tau
    \geq
    \frac{
        m(t)e^{k(t-\tau)}
    }{
        \sqrt{
            2\left[
                \log J_{\max}+k(t-\tau)
            \right]
        }
    }.
    \label{eq:sf_fidelity_lower_bound}
\end{equation}
\end{vsfcorollary}
Thus, within this model,  keeping the
physical-state sensitivity bounded by a threshold imposes a strictly positive lower bound on
the fidelity error.

\begin{tcolorbox}[
  colback=parchment,
  colframe=sandborder,
  boxrule=0.1pt,
  left=8pt,
  right=8pt,
  top=0.3pt,
  bottom=0.3pt
]
\centering
Proposition \ref{prop_motivate} reveals a limitation in  SF:  the physical state must simultaneously track the shared trajectory and encode  future branch information, so near a bifurcation improved physical fidelity necessarily causes branch selection to be more sensitive to state perturbations.
\end{tcolorbox}

\Eqref{eq:bifurcation_fidelity_sensitivity} quantifies the resulting
fidelity--sensitivity tradeoff.

\paragraph{Why introducing a latent model?} While latent variable models are more expressive and can improve performance in general,
the preceding example gives a clear motivation to separate the stochastic branch
information from the physical state, motivating the use of latent SF models.
In the following, we show that there exists such a latent model, without making
a claim about the learned model itself. 
Let $Z\sim\mathcal N(0,1)$ be sampled once for the
trajectory, define
\begin{equation}
b(Z)
=
\begin{cases}
+1, & Z\geq0,\\
-1, & Z<0,
\end{cases}
\label{eq:bifurcation_ideal_branch_code}
\end{equation}
and use the latent-conditioned physical-time field
\begin{equation}
v_{\mathrm{ideal}}(x,t,Z)
=
-kx+b(Z)q(t)
=
b(Z)\dot m(t)
-
k\bigl(x-b(Z)m(t)\bigr),
\label{eq:bifurcation_ideal_field}
\end{equation}
where $q(t)=\dot m(t)+km(t)$ as above.
Thus, conditional on the latent branch code $b(Z)$,
$v_{\mathrm{ideal}}$ has exactly the same form as the branch-conditioned
SF field in \eqref{eq:bifurcation_sf_branch_field}; the difference is that
the branch information is now supplied by the latent variable rather than
having to be inferred from the physical state.

Starting from the  observed state $X_0=0$, this field gives
\begin{equation}
X_t=b(Z)m(t),
\label{eq:bifurcation_ideal_path}
\end{equation}
and hence realizes the target conditional path law
\begin{equation}
\operatorname{Law}(X_\cdot\mid H=h)
=
\frac12\delta_{\xi_+}
+
\frac12\delta_{\xi_-}.
\label{eq:bifurcation_ideal_path_law}
\end{equation}
In particular,
$W_2\left(
P_{\tau}^{\mathrm{ideal}},
P_{\tau}^{\mathrm{data}}
\right)
=0.$
Furthermore,
\begin{equation}
\partial_xv_{\mathrm{ideal}}(x,t,Z)=-k,
\end{equation}
so, for fixed $Z$,
\begin{equation}
\frac{\partial
\Phi_{\tau,t}^{\mathrm{ideal}}(x,Z)}
{\partial x}
=
e^{-k(t-\tau)}.
\label{eq:bifurcation_ideal_sensitivity}
\end{equation}

Indeed, define
$E_t=X_t-b(Z)m(t).$
Using \eqref{eq:bifurcation_ideal_field},
\begin{align}
\dot E_t
&=
\dot X_t-b(Z)\dot m(t)
=
-k\bigl(X_t-b(Z)m(t)\bigr)
=
-kE_t.
\end{align}
Since $X_0=0$ and $m(0)=0$, we have $E_0=0$, and therefore
$E_t=0$ for every $t$, which proves
\eqref{eq:bifurcation_ideal_path}. Since
$\mathbb P(b(Z)=+1)=\mathbb P(b(Z)=-1)=1/2$, this proves
\eqref{eq:bifurcation_ideal_path_law}; since $m(\tau)=0$, we have $W_2\left(
P_{\tau}^{\mathrm{ideal}},
P_{\tau}^{\mathrm{data}}
\right)
=0$.

More generally, for fixed $Z$, solving
\eqref{eq:bifurcation_ideal_field} from time $\tau$ with
$X_\tau=x$ gives
\begin{equation}
\Phi_{\tau,t}^{\mathrm{ideal}}(x,Z)
=
e^{-k(t-\tau)}x+b(Z)m(t),
\end{equation}
where we used $m(\tau)=0$. Differentiating with respect to $x$
gives \eqref{eq:bifurcation_ideal_sensitivity}.
This construction shows what becomes possible once branch information
need not be stored in the physical coordinate: the physical state can remain
on the common stem, while the latent variable distinguishes the future
trajectories.

\paragraph{Connection to rollout robustness.}
The sensitivity calculation also gives a direct local mechanism to understand rollout
error amplification. If a small physical-state perturbation $\delta$ is
introduced at the bifurcation, then standard SF satisfies
\begin{equation}
\Phi_{\tau,t}^{\mathrm{SF}}(\delta)
-
\Phi_{\tau,t}^{\mathrm{SF}}(0)
=
J_{\tau,t}^{\mathrm{SF}}\delta
+
o(\delta),
\label{eq:bifurcation_sf_perturbation}
\end{equation}
whereas the ideal latent construction satisfies, for fixed $Z$,
\begin{equation}
\Phi_{\tau,t}^{\mathrm{ideal}}(x+\delta,Z)
-
\Phi_{\tau,t}^{\mathrm{ideal}}(x,Z)
=
e^{-k(t-\tau)}\delta.
\label{eq:bifurcation_ideal_perturbation}
\end{equation}
Thus, in this example the same mechanism behind the 
fidelity-sensitivity tradeoff also provides a mechanism by which standard
non-latent SF can amplify physical-state rollout errors near a branch
separator. By contrast, once the branch information is encoded by the
latent variable in the idealized construction, physical-state perturbations
are governed only by the original stabilizing dynamics and are contracted
at rate $e^{-k(t-\tau)}$.

\paragraph{This motivates VSF.}
The proposition and discussions above identify a bottleneck of
standard non-latent SF, and the idealized latent construction shows
how that bottleneck can in principle be removed by carrying future-specific
information outside the physical state.
This observation motivates VSF. Rather than requiring the physical state alone
to carry all residual uncertainty needed for later bifurcation, VSF conditions
the physical-time velocity field on an additional latent variable sampled once
per forecast. 
\begin{tcolorbox}[
  colback=parchment,
  colframe=sandborder,
  boxrule=0.1pt,
  left=8pt,
  right=8pt,
  top=0.3pt,
  bottom=0.3pt
]
\centering
The variational  framework of VSF provides a principled way to associate
observed future dynamics with {\it learned latent codes} during training, while the learned prior
provides the corresponding randomness at inference. 
\end{tcolorbox}
The goal is therefore to
learn a distribution of coherent physical-time dynamics in which distinct
future continuations can be represented without requiring increasingly
sensitive separation in the physical state itself.

\clearpage
\section{A variational objective for clean streaming trajectories}
\label{sec:clean_velocity_elbo}

\begin{vsfproposition}[Conditional-velocity evidence lower bound]
\label{prop:velocity_elbo}
Fix a training tuple $(\vu_s,\vx_s,s,\vh)$ constructed from the physical-time
path in Eqs.~\ref{eq:physical_time_interpolation}--
\ref{eq:physical_time_velocity}.  Let $p(\vz)$ be the latent prior, 
$p_{\vtheta}(\vu_s\mid\vx_s,s,\vh,\vz)$ be a conditional velocity
likelihood, and $q_{\vphi}(\vz\mid\vu_s,\vx_s,s,\vh)$ be a
variational posterior.  Define the conditional velocity evidence by
\begin{equation}
  p_{\vtheta}(\vu_s\mid\vx_s,s,\vh)
  :=
  \int
  p_{\vtheta}(\vu_s\mid\vx_s,s,\vh,\vz)
  p(\vz)\,d\vz.
  \label{eq:conditional_velocity_evidence}
\end{equation}
Assume that the displayed quantities are finite, that the evidence is
positive, and that the variational posterior is absolutely continuous with
respect to the exact model posterior.
Then
\begin{align}
  \log p_{\vtheta}(\vu_s\mid\vx_s,s,\vh)
  &\geq
  \mathcal E_{\vtheta,\vphi}(\vu_s,\vx_s,s,\vh)
  \notag\\
  &:={}
  \E_{q_{\vphi}}
  \!\left[
    \log p_{\vtheta}(\vu_s\mid\vx_s,s,\vh,\vz)
  \right]
  \notag
  -
  \KL\!\left(
    q_{\vphi}(\vz\mid\vu_s,\vx_s,s,\vh)
    \,\Vert\,
    p(\vz)
  \right).
  \label{eq:pointwise_velocity_elbo}
\end{align}
\end{vsfproposition}

\paragraph{Proof.}
Inserting the variational posterior into the conditional evidence gives
\begin{align*}
  \log p_{\vtheta}(\vu_s\mid\vx_s,s,\vh)
  &={}
  \log\int
  q_{\vphi}(\vz\mid\vu_s,\vx_s,s,\vh)
  \frac{
    p_{\vtheta}(\vu_s\mid\vx_s,s,\vh,\vz)p(\vz)
  }{
    q_{\vphi}(\vz\mid\vu_s,\vx_s,s,\vh)
  }\,d\vz
  \\
  &={}
  \log\E_{q_{\vphi}(\vz\mid\vu_s,\vx_s,s,\vh)}\!\left[
    \frac{
      p_{\vtheta}(\vu_s\mid\vx_s,s,\vh,\vz)p(\vz)
    }{
      q_{\vphi}(\vz\mid\vu_s,\vx_s,s,\vh)
    }
  \right]
  \\
  &\geq
  \E_{q_{\vphi}(\vz\mid\vu_s,\vx_s,s,\vh)}\!\left[
    \log p_{\vtheta}(\vu_s\mid\vx_s,s,\vh,\vz)
    +\log p(\vz)
    -\log q_{\vphi}(\vz\mid\vu_s,\vx_s,s,\vh)
  \right]
  \\
  &={}
  \E_{q_{\vphi}(\vz\mid\vu_s,\vx_s,s,\vh)}\!\left[
    \log p_{\vtheta}(\vu_s\mid\vx_s,s,\vh,\vz)
  \right]
  \\
  &\quad
  -\KL\!\left(
    q_{\vphi}(\vz\mid\vu_s,\vx_s,s,\vh)
    \,\Vert\,
    p(\vz)
  \right)
  \\
  &={}
  \mathcal E_{\vtheta,\vphi}(\vu_s,\vx_s,s,\vh),
\end{align*}
where Jensen's inequality gives the lower bound. 

For completeness, Bayes' rule gives the exact conditional posterior
\begin{equation*}
  p_{\vtheta}(\vz\mid\vu_s,\vx_s,s,\vh)
  =
  \frac{
    p_{\vtheta}(\vu_s\mid\vx_s,s,\vh,\vz)p(\vz)
  }{
    p_{\vtheta}(\vu_s\mid\vx_s,s,\vh)
  }.
\end{equation*}
Therefore,
\begin{align*}
  &\KL\!\left(
    q_{\vphi}(\vz\mid\vu_s,\vx_s,s,\vh)
    \,\Vert\,
    p_{\vtheta}(\vz\mid\vu_s,\vx_s,s,\vh)
  \right)
  \\
  &={}
  \E_{q_{\vphi}(\vz\mid\vu_s,\vx_s,s,\vh)}\!\left[
    \log\frac{
      q_{\vphi}(\vz\mid\vu_s,\vx_s,s,\vh)
    }{
      p_{\vtheta}(\vu_s\mid\vx_s,s,\vh,\vz)p(\vz)
    }
  \right]
  +\log p_{\vtheta}(\vu_s\mid\vx_s,s,\vh)
  \\
  &={}
  \log p_{\vtheta}(\vu_s\mid\vx_s,s,\vh)
  -\mathcal E_{\vtheta,\vphi}(\vu_s,\vx_s,s,\vh),
\end{align*}
which proves the variational gap where $p_{\vtheta}(\vz\mid\vu_s,\vx_s,s,\vh)$ is the exact model
posterior.  Equality holds if and only if the variational and exact posteriors
agree almost everywhere. \hfill$\square$

\paragraph{Equivalent local conditioning.}
For a realized training tuple, $\vu_s$ is the prediction target of the
conditional velocity likelihood
$p_{\vtheta}(\vu_s\mid\vx_s,s,\vh,\vz)$, rather than one of its
conditioning variables. It is provided only to the training-time variational
posterior $q_{\vphi}$. For the piecewise-linear physical-time path and
$s\in I_k := [s_k, s_{k+1})$,
\begin{equation}
  \vu_s
  =
  P\bigl(\vx_{m+k+1}-\vx_{m+k}\bigr).
  \label{eq:local_velocity_endpoints}
\end{equation}
Conversely, because $s$ determines both $k$ and $\alpha_s$, the two endpoints
can be recovered from $(\vu_s,\vx_s,s)$ according to
\begin{equation}
  \vx_{m+k}
  =
  \vx_s-\frac{\alpha_s}{P}\vu_s,
  \qquad
  \vx_{m+k+1}
  =
  \vx_s+\frac{1-\alpha_s}{P}\vu_s.
  \label{eq:recover_local_endpoints}
\end{equation}
Therefore, on the support of the piecewise-linear interpolation and for almost
every $s$,
\begin{equation}
  (\vu_s,\vx_s,s)
  \quad\Longleftrightarrow\quad
  (\vx_{m+k},\vx_{m+k+1},\vx_s,s).
  \label{eq:local_evidence_equivalence}
\end{equation}
The local endpoint representation thus provides the same conditioning
information to the variational posterior as $(\vu_s,\vx_s,s)$. This
reparameterization does not alter the conditioning variables
$(\vx_s,s,\vh)$ of the velocity likelihood. Accordingly, the local
variational posterior can equivalently be parameterized as
\begin{equation}
  q_{\vphi}^{\mathrm{loc}}\!\left(
    \vz
    \mid
    \vx_{m+k},
    \vx_{m+k+1},
    \vx_s,
    s,
    \vh
  \right).
  \label{eq:local_variational_encoder}
\end{equation}

\clearpage

\section{Population Energy Reconstruction and Energy-Distance Control}
\label{sec:population_energy_control}

Fix $(\vx_s,s,\vh)$ and draw
\begin{align*}
  \vu,\vu'
  &\stackrel{\mathrm{iid}}{\sim}
  p_{\mathrm{data}}(\cdot\mid\vx_s,s,\vh),
  \quad
  \vz,\vz'\mid\vu
  \stackrel{\mathrm{iid}}{\sim}
  q_{\vphi}(\cdot\mid\vu,\vx_s,s,\vh),
  \quad
  \widetilde{\vz},\widetilde{\vz}'
  \stackrel{\mathrm{iid}}{\sim}p(\vz),
\end{align*}
where all draws are independent except for the displayed conditioning. Define
\begin{align*}
  \widehat{\vu}
  &:=
  \vv_{\vtheta}(\vx_s,s,\vh,\vz),
  &
  \widehat{\vu}'
  &:=
  \vv_{\vtheta}(\vx_s,s,\vh,\vz'),
  \\
  \widetilde{\vu}
  &:=
  \vv_{\vtheta}(\vx_s,s,\vh,\widetilde{\vz}),
  &
  \widetilde{\vu}'
  &:=
  \vv_{\vtheta}(\vx_s,s,\vh,\widetilde{\vz}').
\end{align*}
For a realized target $\vu$, let
\begin{equation}
  S_\alpha(\vu)
  :=
  \E\!\left[
    \left\|\widehat{\vu}-\vu\right\|_2^\alpha
    \,\middle|\,\vu
  \right]
  -
  \frac12
  \E\!\left[
    \left\|\widehat{\vu}-\widehat{\vu}'\right\|_2^\alpha
    \,\middle|\,\vu
  \right].
  \label{eq:vsf_population_energy_reconstruction}
\end{equation}
This is the population counterpart of the generalized energy reconstruction
in Eq.~\ref{eq:vsf_energy_score}; replacing its expectations with the $M$
posterior samples gives the finite-ensemble V-statistic
$\mathcal L_{\mathrm{ES}}$.

The energy distance between the prior-predictive velocity and the true
conditional velocity is
\begin{equation}
  D_\alpha(\widetilde{\vu},\vu)
  :=
  2\E\left\|\widetilde{\vu}-\vu\right\|_2^\alpha
  -
  \E\left\|\widetilde{\vu}-\widetilde{\vu}'\right\|_2^\alpha
  -
  \E\left\|\vu-\vu'\right\|_2^\alpha.
  \label{eq:vsf_prior_energy_distance}
\end{equation}

Define the aggregate posterior
\begin{equation}
    \overline q_{\vphi}(\vz\mid\vx_s,s,\vh)
    :=
    \E_{\vu\sim
    p_{\mathrm{data}}(\cdot\mid\vx_s,s,\vh)}
    \left[
        q_{\vphi}(\vz\mid\vu,\vx_s,s,\vh)
    \right].
    \label{eq:aggregate_posterior}
\end{equation}

\begin{vsfproposition}[Prior-predictive energy-distance control]
\label{prop:vsf_aggregate_energy_bound}
If $0<\alpha<2$, all expectations above are finite, and
\(
  \overline q_{\vphi}(\vz\mid\vx_s,s,\vh)=p(\vz)
  \quad\text{for almost every }\vz,
\)
then
\begin{equation}
  \frac12D_\alpha(\widetilde{\vu},\vu)
  \leq
  \E_{\vu\sim p_{\mathrm{data}}(\cdot\mid\vx_s,s,\vh)}
  \!\left[S_\alpha(\vu)\right]
  \label{eq:vsf_aggregate_energy_bound}
\end{equation}
\end{vsfproposition}

\paragraph{Proof.}
For $\rho_\alpha(a,b)=\|a-b\|_2^\alpha$ and $0<\alpha<2$, there exist a
Hilbert space $\mathcal H_\alpha$ and a feature map $\psi_\alpha$ such that
\begin{equation*}
  \rho_\alpha(a,b)
  =
  \left\|\psi_\alpha(a)-\psi_\alpha(b)\right\|_{\mathcal H_\alpha}^2
\end{equation*}
\citep{gneiting2007strictly,szekely2013energy}. Define
\begin{equation*}
  m_q(\vu)
  :=
  \E[\psi_\alpha(\widehat{\vu})\mid\vu],
  \qquad
  m_p
  :=
  \E[\psi_\alpha(\widetilde{\vu})].
\end{equation*}
Aggregate posterior--prior alignment and
\(\overline q_{\vphi}(\vz\mid\vx_s,s,\vh)
    :=
    \E_{\vu\sim p_{\mathrm{data}}(\cdot\mid\vx_s,s,\vh)}
    \!\left[
        q_{\vphi}(\vz\mid\vu,\vx_s,s,\vh)
    \right]\) imply

\begin{equation*}
  m_p
  =
  \E_{\vu\sim p_{\mathrm{data}}(\cdot\mid\vx_s,s,\vh)}
  [m_q(\vu)].
\end{equation*}
The energy-score identities and Jensen's inequality now give
\begin{align*}
  \frac12D_\alpha(\widetilde{\vu},\vu)
  &=
  \left\|
    \E_{\vu}
    \!\left[m_q(\vu)-\psi_\alpha(\vu)\right]
  \right\|_{\mathcal H_\alpha}^2
  \\
  &\leq
  \E_{\vu}
  \!\left[
    \left\|m_q(\vu)-\psi_\alpha(\vu)\right\|_{\mathcal H_\alpha}^2
  \right]
  =
  \E_{\vu}[S_\alpha(\vu)],
\end{align*}
where $\vu\sim p_{\mathrm{data}}(\cdot\mid\vx_s,s,\vh)$ in each
expectation. \hfill$\square$

The expected energy reconstruction upper-bounds half the energy distance between the decoded aggregate posterior and the true conditional velocity law. Thus, reducing the reconstruction objective controls a distribution-level discrepancy, rather than only an error in the conditional mean.

\paragraph{Interpretation of aggregate alignment.}
For each fixed context $c=(\vx_s,s,\vh)$, the equality says that averaging
the encodings of true velocities at that context gives the prior. It can hold
in several circumstances:
\begin{itemize}[leftmargin=*,nosep]
  \item \textbf{Ideal inference and model fit.}
  If $q_{\vphi}(\vz\mid\vu,c)$ is the exact model posterior and
  $p_{\vtheta}(\vu\mid c)=p_{\mathrm{data}}(\vu\mid c)$, then Bayes' rule
  gives
  \begin{equation*}
    \overline q_{\vphi}(\vz\mid c)
    =
    \int
    p_{\vtheta}(\vz\mid\vu,c)
    p_{\vtheta}(\vu\mid c)\,d\vu
    =
    p(\vz).
  \end{equation*}
  This is the intended justification for the assumption
  \citep{hoffman2016elbo}.

  \item \textbf{Aggregate-matching training.}
  A penalty on the aggregate posterior can encourage approximate equality,
  although finite-capacity training does not guarantee exact equality
  \citep{tolstikhin2017wasserstein}.

  \item \textbf{Posterior collapse.}
  The encoder can output $q_{\vphi}(\vz\mid\vu,c)=p(\vz)$ for every $\vu$,
  which also satisfies the equality.
\end{itemize}

\clearpage
\section{Training and inference details}

\subsection{Forecasting training and inference}
\label{sec:vsf-forecasting}
\Cref{alg:vsf-training,alg:vsf-inference} summarize the two procedures.
The current state and rolling window
are updated at every step, while $\vh$ is fixed within each $K$-step segment.
Because future observations are unavailable in forecasting, $\vh$ is
recomputed from the generated-state window at each segment boundary.

\noindent
\begin{minipage}[t]{0.485\linewidth}
\vspace{0pt}
\begin{algorithm}[H]
\caption{VSF training}
\label{alg:vsf-training}
\begin{vsfalgorithmic}
\VSFInput Training windows $\mathcal D=\{\vx_{a:T}\}$;
posterior samples $M$; $\alpha$ and $\beta_{\mathrm{KL}}$.
\VSFOutput Trained velocity field $\vv_{\vtheta}$ and posterior $q_{\vphi}$.
\VSFWhile{not converged}
  \VSFState Sample $\vx_{a:T}\sim\mathcal D$ and $s\sim\mathcal U[0,1]$.
  \VSFState Set $\vh\gets \vx_{a:m}$ and construct $(\vx_s,\vu_s)$ by
  \eqref{eq:physical_time_interpolation}--\eqref{eq:physical_time_velocity}.
  \VSFState Draw $\vz^{(1:M)}\stackrel{\mathrm{iid}}{\sim}
  q_{\vphi}(\vz\mid\vu_s,\vx_s,s,\vh)$.
  \VSFState Evaluate in one batch: $\widehat{\vu}_s^{(1:M)}\gets
  \vv_{\vtheta}(\vx_s,s,\vh,\vz^{(1:M)})$.
  \VSFState Evaluate $\mathcal L_{\mathrm{VSF}}$ by
  \eqref{eq:vsf_training_objective}.
  \VSFState Update $(\vtheta,\vphi)$ using
  $\nabla_{\vtheta,\vphi}\mathcal L_{\mathrm{VSF}}$.
\VSFEndWhile
\end{vsfalgorithmic}
\end{algorithm}
\end{minipage}\hfill
\begin{minipage}[t]{0.485\linewidth}
\vspace{0pt}
\begin{algorithm}[H]
\caption{VSF inference}
\label{alg:vsf-inference}
\begin{vsfalgorithmic}
\VSFInput $\vv_{\vtheta}$, $\vx_{a:m}$; integration
step $\Delta t=1/P$; segment length $K$; forecast endpoint $T$.
\VSFOutput Forecast $\widehat{\vx}_{m+1:T}$.
\VSFState Draw $\vz\sim p(\vz)$ once for the complete rollout.
\VSFState $\mathcal W\gets\vx_{a:m}$, $\widehat{\vx}\gets\vx_m$, and
$\vh\gets \mathcal W$.
\VSFWhile{the forecast is incomplete}
  \VSFState Set $s\gets0$ and keep $\vh$ fixed below.
  \VSFFor{up to $K$ steps while the forecast is incomplete}
    \VSFState $\widehat{\vx}\gets\widehat{\vx}
    +\Delta t\,\vv_{\vtheta}(\widehat{\vx},s,\vh,\vz)$.
    \VSFState $s\gets s+\Delta t$; emit $\widehat{\vx}$ as the next state.
    \VSFState $\mathcal W\gets
    \operatorname{Update}(\mathcal W,\widehat{\vx})$.
  \VSFEndFor
  \VSFState $\vh\gets \mathcal W$ \emph{(refresh from generated states)}.
\VSFEndWhile
\end{vsfalgorithmic}
\end{algorithm}
\end{minipage}\par

\smallskip
\noindent
Here $\operatorname{Update}$ is a fixed-length queue: it enqueues the generated state and dequeues the oldest state
when the window is full. The integration step
$\Delta t$ advances the normalized forecast time $s$. The history encoding
$\vh$ is refreshed after each segment, while $\vz$ is fixed throughout
the complete rollout.

\medskip
\noindent

\clearpage
\section{World-model VSF predictor:  training and inference}
\label{sec:wm-vsf-algorithms}

The world-model implementation specializes the preceding procedures to the
LeWM latent space. Let $\mathcal E$ denote the frozen JEPA representation map
(visual encoder followed by its projection head), and let $N$ be the
causal-context length. Write $\vx_k=\mathcal E(\vo_k)$, with $\va_k$ aligned
with the transition from $\vx_k$ to $\vx_{k+1}$. A training clip
contains $N+1$ observations and $N$ aligned actions. The causal token window
plays the role of history and is refreshed after every imagined transition;
no separate representation $\vh$ is cached. Only the VSF predictor and its
trajectory posterior are trained. Inference below uses the single-frame
planning initialization evaluated in our experiments.

For each training clip, draw one $\rho\sim\mathcal U[0,1]$ and, for
$k=0,\ldots,N-1$, define the teacher-forced tuples
\[
  s_k=\frac{k+\rho}{N},\qquad
  \vx_{s_k}=(1-\rho)\vx_k+\rho\vx_{k+1},\qquad
  \vu_{s_k}=N(\vx_{k+1}-\vx_k).
\]

Following LeWM~\citep{maes2026leworldmodel}, training uses teacher forcing with
temporal causal masking. Every $\vx_{s_k}$ is constructed from the observed pair
$(\vx_k,\vx_{k+1})$, and the predictor never consumes an earlier prediction. In
contrast, inference is open-loop and feeds each generated latent back into the
rolling causal context. Define
$\widehat{\vu}_{s_k}^{(j)}:=[\vv_{\vtheta}(\vx_{\vs},\vs,
\va_{0:N-1},\vz^{(j)})]_k$. Causal masking restricts this prediction to the
prefix through $k$. Averaging the energy reconstruction in
\eqref{eq:vsf_energy_score} over the $N$ teacher-forced transitions gives
\begin{equation}
\begin{aligned}
\mathcal L_{\mathrm{WM\text{-}VSF}}^{\mathrm{TF}}
={}&\frac{1}{NM}\sum_{k=0}^{N-1}\sum_{j=1}^{M}
\left\|\widehat{\vu}_{s_k}^{(j)}-\vu_{s_k}\right\|_2^\alpha \\
&-\frac{1}{2NM^2}\sum_{k=0}^{N-1}\sum_{i,j=1}^{M}
\left\|\widehat{\vu}_{s_k}^{(i)}-
\widehat{\vu}_{s_k}^{(j)}\right\|_2^\alpha \\
&+\beta_{\mathrm{KL}}\KL\!\left(
q_{\vphi}(\vz\mid\vx_{0:N},\va_{0:N-1},\vx_{\vs},\vs)
\,\Vert\,p(\vz)\right),
\qquad 0<\alpha<2.
\end{aligned}
\label{eq:wm-vsf-teacher-forcing-loss}
\end{equation}

\medskip
\noindent
\begin{minipage}[t]{0.485\linewidth}
\vspace{0pt}
\begin{algorithm}[H]
\caption{World-model VSF training}
\label{alg:wm-vsf-training}
\begin{vsfalgorithmic}
\VSFInput Offline clips
$\mathcal D_{\mathrm{WM}}=\{(\vo_{0:N},\va_{0:N-1})\}$; frozen
$\mathcal E$; posterior samples $M$;
$\alpha$ and $\beta_{\mathrm{KL}}$.
\VSFOutput Trained predictor $\vv_{\vtheta}$ and posterior $q_{\vphi}$
\VSFWhile{not converged}
  \VSFState Sample a minibatch from $\mathcal D_{\mathrm{WM}}$ and encode
  $\vx_{0:N}\gets\mathcal E(\vo_{0:N})$.
  \VSFState Construct $(\vs,\vx_{\vs},\vu_{\vs})$ from the
  teacher-forced tuples defined above.
  \VSFState Form
  $q_{\vphi}(\vz\mid\vx_{0:N},\va_{0:N-1},\vx_{\vs},\vs)$ and draw
  $\vz^{(1:M)}$ by batched reparameterization.
  \VSFState In one teacher-forced causal batched call, evaluate
  $\widehat{\vu}_{\vs}^{(1:M)}\gets
  \vv_{\vtheta}(\vx_{\vs},\vs,\va_{0:N-1},\vz^{(1:M)})$.
  \VSFState Evaluate $\mathcal L_{\mathrm{WM\text{-}VSF}}^{\mathrm{TF}}$
  by \eqref{eq:wm-vsf-teacher-forcing-loss} and update $(\vtheta,\vphi)$.
\VSFEndWhile
\end{vsfalgorithmic}
\end{algorithm}
\end{minipage}\hfill
\begin{minipage}[t]{0.485\linewidth}
\vspace{0pt}
\begin{algorithm}[H]
\caption{World-model VSF inference}
\label{alg:wm-vsf-inference}
\begin{vsfalgorithmic}
\VSFInput Current observation $\vo_m$; candidate actions
$\va_{m:T-1}$; frozen $\mathcal E$; trained
$\vv_{\vtheta}$; and step $\Delta t=1/N$.
\VSFOutput Imagined latent trajectory $\widehat{\vx}_{m+1:T}$.
\VSFState Draw $\vz\sim p(\vz)$ for the complete rollout.
\VSFState Set $s\gets0$ and $\widehat{\vx}_s\gets\mathcal E(\vo_m)$;
initialize rolling sequences
$\widehat{\vx}_{\mathrm{win}},\va_{\mathrm{win}}\gets\emptyset$.
\VSFWhile{the candidate plan has an unused action}
  \VSFState Take the next action $\va$; append $\widehat{\vx}_s$ to
  $\widehat{\vx}_{\mathrm{win}}$ and $\va$ to $\va_{\mathrm{win}}$; retain
  their $N$ most recent entries.
  \VSFState Set $L\gets|\va_{\mathrm{win}}|$ and
  $\vs\gets(0,\Delta t,\ldots,(L-1)\Delta t)$.
  \VSFState $\widehat{\vu}_s\gets\left[
  \vv_{\vtheta}(\widehat{\vx}_{\mathrm{win}},\vs,
  \va_{\mathrm{win}},\vz)\right]_{L-1}$.
  \VSFState $\widehat{\vx}_{s+\Delta t}\gets
  \widehat{\vx}_s+\Delta t\,\widehat{\vu}_s$; emit it and set
  $s\gets s+\Delta t$.
\VSFEndWhile
\end{vsfalgorithmic}
\end{algorithm}
\end{minipage}\par

\medskip
\noindent
\clearpage
\section{Evaluation metrics}
\label{sec:metrics}

Let $c_b$, $b=1,\ldots,B$, denote the test conditions. For forecast time
$t=1,\ldots,L$, let $\mathbf y_{b,t}\in\mathbb{R}^{d}$ denote the state, and
write $\mathbf y_b=(\mathbf y_{b,1},\ldots,\mathbf y_{b,L})\in
\mathbb{R}^{Ld}$ for the complete future trajectory. For condition $c_b$, let
\[
P_b := P(\,\cdot\mid c_b),
\qquad
Q_{\theta,b} := Q_\theta(\,\cdot\mid c_b)
\]
denote the reference and learned conditional distributions, respectively.
The model produces an ensemble
$\{\widehat{\mathbf y}_b^{(m)}\}_{m=1}^{M}$ from $Q_{\theta,b}$, with
$\widehat{\mathbf y}_{b,t}^{(m)}\in\mathbb R^d$ denoting the state at time
$t$ in ensemble member $m$.
All reported metrics are averaged over test conditions, and lower values are
better. The definitions below give the full-horizon scores reported in the
main tables; their relationship to the per-timestep rollout curves is described below.

\subsection{Deterministic ground truth}

For deterministic systems, $P_b=\delta_{\mathbf y_b}$ is concentrated at the
single ground-truth future $\mathbf y_b\in\mathbb{R}^{Ld}$. The metrics in
this subsection are pointwise in forecast time: they evaluate time-local
states and then aggregate over forecast times, without comparing the joint
temporal distribution.

\paragraph{Continuous ranked probability score (CRPS).}
CRPS is a proper scoring rule for comparing a scalar predictive cumulative
distribution function $F$ with an observation $y$
\citep{gneiting2007strictly, waghmare2025proper}.  It is defined as
\begin{equation}
  \operatorname{CRPS}(F,y)
  =
  \int_{-\infty}^{\infty}
  \left[F(a)-\mathbf{1}\{a\geq y\}\right]^2\,\mathrm{d}a.
  \label{eq:crps-cdf}
\end{equation}
Equivalently, if $X$ and $X'$ are independent draws from $F$, then
\begin{equation}
  \operatorname{CRPS}(F,y)
  =
  \mathbb{E}|X-y|
  -\frac{1}{2}\mathbb{E}|X-X'|.
  \label{eq:crps-expectation}
\end{equation}
We apply this score coordinatewise to the state at each forecast time. Its
finite-ensemble estimate, averaged over conditions, forecast times, and state
coordinates, is
\begin{equation}
  \begin{aligned}
    \widehat{\operatorname{CRPS}}
    ={}&
    \frac{1}{BLd}\sum_{b=1}^{B}\sum_{t=1}^{L}
    \left[
      \frac{1}{M}\sum_{m=1}^{M}
      \left\|\widehat{\mathbf y}_{b,t}^{(m)}-\mathbf y_{b,t}\right\|_1
    \right.\\[-1mm]
    &\left.\hspace{30mm}
      -\frac{1}{2M^2}\sum_{m=1}^{M}\sum_{m'=1}^{M}
      \left\|\widehat{\mathbf y}_{b,t}^{(m)}-
      \widehat{\mathbf y}_{b,t}^{(m')}\right\|_1
    \right].
  \end{aligned}
  \label{eq:crps-ensemble}
\end{equation}
The $\ell_1$ norms in Eq.~\ref{eq:crps-ensemble} are compact notation for
summing scalar CRPS terms over the $d$ state coordinates; they do not make
CRPS a joint multivariate or trajectory-level score.

\paragraph{Root mean squared error (RMSE) and mean absolute error (MAE).}
Both metrics use the time-local ensemble mean
$\overline{\mathbf y}_{b,t}=M^{-1}\sum_{m=1}^{M}
\widehat{\mathbf y}_{b,t}^{(m)}$ as the point forecast:
\begin{equation}
  \begin{aligned}
    \operatorname{RMSE}
    &=
    \left[
      \frac{1}{BLd}\sum_{b=1}^{B}\sum_{t=1}^{L}
      \left\|\overline{\mathbf y}_{b,t}-\mathbf y_{b,t}\right\|_2^2
    \right]^{1/2},\\
    \operatorname{MAE}
    &=
    \frac{1}{BLd}\sum_{b=1}^{B}\sum_{t=1}^{L}
    \left\|\overline{\mathbf y}_{b,t}-\mathbf y_{b,t}\right\|_1.
  \end{aligned}
  \label{eq:rmse-mae}
\end{equation}

\subsection{Stochastic ground truth}

For stochastic systems, the simulator provides a reference ensemble
$\{\mathbf y_b^{(r)}\}_{r=1}^{R}$ from $P_b$, which we compare with the model
ensemble from $Q_{\theta,b}$. For forecast time
$t$, let $P_{b,t}$ and $Q_{\theta,b,t}$ denote the $d$-dimensional marginal
laws of the state at time $t$ under $P_b$ and $Q_{\theta,b}$, respectively. The metrics below define the full-horizon
scores reported in the main tables. For the rollout plots, we evaluate each
metric separately at each forecast time: CME and CSE use the time-$t$ moments
in their relative-error ratios, while $W_1$ and ED compare the time-$t$
marginal ensembles.

\paragraph{Conditional mean error (CME) and
conditional standard-deviation error (CSE).}
Let $\boldsymbol{\mu}_{P,b,t}$ and $\boldsymbol{\mu}_{Q,b,t}$ denote the mean
vectors of $P_{b,t}$ and $Q_{\theta,b,t}$, and let
$\boldsymbol{\sigma}_{P,b,t}$ and $\boldsymbol{\sigma}_{Q,b,t}$ denote their
coordinatewise standard-deviation vectors. The full-horizon scores reported in the main tables aggregate
the resulting marginal-moment errors over all forecast times:
\begin{equation}
  \begin{aligned}
    \operatorname{CME}
    &=
    \frac{1}{B}\sum_{b=1}^{B}
    \frac{
      \left[
        \sum_{t=1}^{L}
        \left\|\boldsymbol{\mu}_{P,b,t}-\boldsymbol{\mu}_{Q,b,t}\right\|_2^2
      \right]^{1/2}
    }{
      \left[
        \sum_{t=1}^{L}\left\|\boldsymbol{\mu}_{P,b,t}\right\|_2^2
      \right]^{1/2}
    },
    \\
    \operatorname{CSE}
    &=
    \frac{1}{B}\sum_{b=1}^{B}
    \frac{
      \left[
        \sum_{t=1}^{L}
        \left\|\boldsymbol{\sigma}_{P,b,t}-\boldsymbol{\sigma}_{Q,b,t}\right\|_2^2
      \right]^{1/2}
    }{
      \left[
        \sum_{t=1}^{L}\left\|\boldsymbol{\sigma}_{P,b,t}\right\|_2^2
      \right]^{1/2}
    }.
  \end{aligned}
  \label{eq:cme-cse}
\end{equation}
The moments are estimated from the respective ensembles using population
standard deviations. Thus, CME and CSE use the complete horizon in their
final norm, but remain pointwise-marginal rather than trajectory-level
distribution metrics.

By contrast, the $W_1$ and ED metrics below are trajectory-level: each
ensemble member is retained as one complete vector in $\mathbb R^{Ld}$, so
their comparisons are sensitive to cross-time dependence.

\paragraph{First-order Wasserstein distance ($W_1$).}
Following its standard use in generative modeling
\citep{arjovsky2017wasserstein}, the population distance is
\begin{equation}
  W_1(P_b,Q_{\theta,b})
  =
  \inf_{\gamma\in\Pi(P_b,Q_{\theta,b})}
  \mathbb{E}_{(\mathbf Y,\widehat{\mathbf Y})\sim\gamma}
  \left[\left\|\mathbf Y-\widehat{\mathbf Y}\right\|_2\right],
  \label{eq:wasserstein-one}
\end{equation}
where $\Pi(P_b,Q_{\theta,b})$ is the set of couplings with marginals $P_b$
and $Q_{\theta,b}$.  For the two empirical ensembles, define the cost matrix
$C^{(b)}_{rm}=\|\mathbf y_b^{(r)}-\widehat{\mathbf y}_b^{(m)}\|_2$.
We compute the entropically regularized transport plan
\citep{cuturi2013sinkhorn}
\begin{equation}
  \gamma_{b,\varepsilon}^{\star}
  \in
  \argmin_{\gamma\in\mathcal{U}_{R,M}}
  \left\langle\gamma,C^{(b)}\right\rangle
  -\varepsilon H(\gamma),
  \qquad
  \widehat W_{1,\varepsilon}
  =
  \frac{1}{B}\sum_{b=1}^{B}
  \left\langle\gamma_{b,\varepsilon}^{\star},C^{(b)}\right\rangle,
  \label{eq:sinkhorn-wasserstein}
\end{equation}
where $\mathcal{U}_{R,M}$ contains nonnegative transport plans with uniform
marginals and $H$ is entropy.  We use $\varepsilon=0.1$ and 100 Sinkhorn
iterations.

\paragraph{Energy distance (ED).}
For independent $\mathbf Y,\mathbf Y'\sim P_b$ and
$\widehat{\mathbf Y},\widehat{\mathbf Y}'\sim Q_{\theta,b}$, energy distance is
defined by \citep{szekely2013energy}
\begin{equation}
  \operatorname{ED}(P_b,Q_{\theta,b})
  =
  2\,\mathbb{E}\left\|\mathbf Y-\widehat{\mathbf Y}\right\|_2
  -\mathbb{E}\left\|\mathbf Y-\mathbf Y'\right\|_2
  -\mathbb{E}\left\|\widehat{\mathbf Y}-\widehat{\mathbf Y}'\right\|_2.
  \label{eq:energy-distance}
\end{equation}
We estimate this quantity with the unbiased finite-ensemble statistic
\begin{align}
  \widehat{\operatorname{ED}}_b
  ={}&
  \frac{2}{RM}\sum_{r=1}^{R}\sum_{m=1}^{M}
  \left\|\mathbf y_b^{(r)}-\widehat{\mathbf y}_b^{(m)}\right\|_2
  \nonumber\\
  &-
  \frac{1}{R(R-1)}
  \sum_{\substack{r,r'=1\\r\neq r'}}^{R}
  \left\|\mathbf y_b^{(r)}-\mathbf y_b^{(r')}\right\|_2
  \nonumber\\
  &-
  \frac{1}{M(M-1)}
  \sum_{\substack{m,m'=1\\m\neq m'}}^{M}
  \left\|\widehat{\mathbf y}_b^{(m)}-
  \widehat{\mathbf y}_b^{(m')}\right\|_2,
  \label{eq:energy-distance-ensemble}
\end{align}
and report
$\widehat{\operatorname{ED}}=B^{-1}\sum_{b=1}^{B}
\widehat{\operatorname{ED}}_b$.
Unlike the coordinatewise CRPS, both $W_1$ and ED use Euclidean distances
between complete trajectories and therefore evaluate the joint temporal
distribution.

\clearpage
\section{Dataset Statistics}
\label{sec:dataset-statistics}

We evaluate five forecasting datasets: three deterministic systems and two
conditional parameter-mixture variants.  Table~\ref{tab:forecasting-datasets}
summarizes the data used in the reported experiments.   We integrate every
system with the adaptive DOP853 solver in 64-bit floating-point precision,
using relative and absolute error tolerances of $10^{-9}$.  We store the
resulting trajectories in 32-bit floating-point precision for training.  The $\Delta t$ values in
Table~\ref{tab:forecasting-datasets} specify the uniform output-sampling
intervals; the solver selects its internal step sizes adaptively.

\begin{table}[H]
\centering
\small
\setlength{\tabcolsep}{3.5pt}
\renewcommand{\arraystretch}{1.08}
\resizebox{\linewidth}{!}{%
\begin{tabular}{@{}llrrrrrr@{}}
\toprule
Dataset & Regime & \shortstack{State\\dimension}
& \shortstack{Initial\\conditions}
& \shortstack{Trajectories per\\initial condition}
& \shortstack{Time\\steps} & $\Delta t$
& \shortstack{History\\length} \\
\midrule
Double pendulum & Deterministic & 8 & 200 & 1   & 1,200 & 0.01 & 4 \\
R\"ossler       & Deterministic & 3 & 200 & 1   & 1,500 & 0.02 & 10 \\
Lorenz          & Deterministic & 3 & 200 & 1   & 3,000 & 0.01 & 5 \\
R\"ossler       & Parameter mixture & 3 & 40 & 100 & 1,500 & 0.02 & 10 \\
Lorenz          & Parameter mixture & 3 & 40 & 100 & 3,000 & 0.01 & 5 \\
\bottomrule
\end{tabular}%
}
\caption{Forecasting dataset statistics.  For deterministic datasets, each
sampled initial condition produces one trajectory.  For parameter-mixture
datasets, each initial condition is paired with 100 independently sampled
dynamical parameters, producing 100 trajectories from that same initial
state. 
We verify that these trajectories remain closely aligned for at least the full conditioning history: across all within-group trajectory pairs and conditioning time points, the mean pointwise RMSE is $0.008$ for Rössler and $0.047$ for Lorenz. Their subsequent futures are allowed to diverge.
The integration intervals are $[0,12]$ for double pendulum and
$[0,30]$ for R\"ossler and Lorenz.}
\label{tab:forecasting-datasets}
\end{table}

\subsection{Deterministic Datasets}
\label{sec:deterministic-datasets}

\paragraph{Double pendulum.}
We use only the Cartesian state in the forecasting experiments.  Let
$(x_i,y_i)$ denote the position of mass $i$.  The constrained Cartesian
dynamics are
\[
\begin{array}{@{}l@{\qquad\qquad}l@{}}
  m_1\ddot{x}_1=-\lambda_1x_1+\lambda_2(x_2-x_1),
  &m_2\ddot{x}_2=-\lambda_2(x_2-x_1),\\
  m_1\ddot{y}_1=-m_1g-\lambda_1y_1+\lambda_2(y_2-y_1),
  &m_2\ddot{y}_2=-m_2g-\lambda_2(y_2-y_1),\\
  x_1^2+y_1^2=l_1^2,
  &(x_2-x_1)^2+(y_2-y_1)^2=l_2^2.
\end{array}
\]
where $\lambda_1$ and $\lambda_2$ enforce the rod-length constraints.  We set
$m_1=m_2=l_1=l_2=1$ and $g=9.81$.  For each of the 200 trajectories, we
independently sample the initial rod orientations from
$\mathcal U[-\pi/2,\pi/2]$ and their angular rates from
$\mathcal U[-0.5,0.5]$, then convert the initial condition to the Cartesian
state
$\vx=(x_1,y_1,x_2,y_2,v_{x_1},v_{y_1},v_{x_2},v_{y_2})\in\mathbb R^8$.

\paragraph{R\"ossler system.}
The R\"ossler dynamics are
\begin{equation}
  \dot{x}=-y-z,\qquad
  \dot{y}=x+ay,\qquad
  \dot{z}=b+z(x-c).
  \label{eq:rossler-dynamics}
\end{equation}
We set $a=b=0.2$ and $c=5.7$.  To generate each of the 200 initial
conditions, we select a center uniformly from
$\{-1,1\}^3$; the initial-condition perturbation has standard deviation
$0.01$.

\paragraph{Lorenz system.}
The Lorenz dynamics are
\begin{equation}
  \dot{x}=\sigma(y-x),\qquad
  \dot{y}=\rho x-y-xz,\qquad
  \dot{z}=xy-\beta z.
  \label{eq:lorenz-dynamics}
\end{equation}
We use $\sigma=10$, $\beta=8/3$, and $\rho=24.74$.  Each of the 200
initial conditions is centered at $(1,1,1)$, with perturbation standard
deviation $0.01$.

\subsection{Stochastic Datasets}
\label{sec:stochastic-datasets}

\paragraph{Stochastic data-generation.}
These benchmarks contain uncertainty over a trajectory-level dynamical
parameter rather than additive process noise.  For each initial condition
$i$, we sample one state $\vx_0^{(i)}$ and hold it fixed while generating
$100$ trajectories.  For each trajectory $j$, we independently draw a
parameter $\eta_{ij}$ and integrate the resulting deterministic system:
\begin{equation}
  \eta_{ij}\sim\mathcal U(\eta_{\min},\eta_{\max}),\qquad
  \vx_{ij}(t)
  =\operatorname{ODESolve}\!\left(f_{\eta_{ij}},\vx_0^{(i)},t\right).
  \label{eq:stochastic-dataset-generation}
\end{equation}
The parameter is withheld from the forecaster, so a shared observed context
corresponds to a distribution of possible futures.  We split the 40 initial
conditions, rather than the resulting 4,000 trajectories, into 70\%/15\%/15\%
train/validation/test partitions.  For evaluation, we condition on each held-out initial condition and use all 100 corresponding futures as samples from its reference conditional distribution.

\paragraph{Stochastic R\"ossler system.}
We use Eq.~\ref{eq:rossler-dynamics} with $a=b=0.2$ and independently sample
$c_{ij}\sim\mathcal U[5.2,6.2]$ for every trajectory.  Each initial condition
is obtained by selecting a center uniformly from
$\{-1,1\}^3$, with perturbation standard deviation $0.01$.

\paragraph{Stochastic Lorenz system.}
We use Eq.~\ref{eq:lorenz-dynamics} with $\sigma=10$ and $\beta=8/3$, and
independently sample $\rho_{ij}\sim\mathcal U[23.74,25.74]$ for every
trajectory.  Each initial condition is centered at $(1,1,1)$, with
perturbation standard deviation $0.01$.

\subsection{World-Model Datasets}
\label{sec:world-model-datasets}

We use the four offline visual-control datasets in LeWM~\citep{maes2026leworldmodel}. Trajectories contain $224\times224$ RGB images, continuous actions, and auxiliary simulator state. Following LeWM, we sample images every five environment steps and form four-frame clips with a five-action block per transition. The predictor uses only frozen LeWM image embeddings and actions. Auxiliary state is used only to restore evaluation starts and goals and compute success.

\paragraph{PushT.}
In this continuous 2D task, a blue point agent pushes a T-shaped block to a demonstrated future configuration~\citep{chi2025diffusion}. Block interactions are limited to pushing. The goal specifies both pusher and block poses. Following LeWM, we reuse noisy DINO-WM expert replay~\citep{zhou2025dinowm} comprising 20,000 episodes averaging 196 steps.

\paragraph{Cube.}
OGBench-Cube is the single-object variant of the OGBench manipulation
benchmark~\citep{park2025ogbench}.  A UR5e arm with a parallel-jaw gripper is
controlled through end-effector displacement, yaw, and gripper commands to
pick up a cube and place it at a target location.   The dataset contains 10,000 200-step
episodes collected with the benchmark's scripted collection heuristic.  Success depends on the cube center reaching its target, regardless of cube orientation.

\paragraph{TwoRoom.}
A red point agent starts at a random position in one room and navigates to a sampled target in the other through the single doorway~\citep{sobal2025stress}. The dataset contains 10,000 episodes averaging 92 steps, generated by a noisy doorway-then-target heuristic. 

\paragraph{Reacher.}
Reacher is based on the DeepMind Control Suite planar two-link
arm~\citep{tassa2018deepmind}.  We use the DINO-WM/LeWM \emph{qpos-match}
variant, in which the goal is a complete future joint configuration.   The dataset contains 10,000 200-step episodes collected with a
Soft Actor--Critic policy.

\clearpage
\section{Experimental Setup}
\label{sec:experimental-setup}

\subsection{Forecasting}
\label{sec:forecasting-setup}

\paragraph{Baseline models.}
We compare VSF with eight baselines: Block diffusion, which jointly denoises
a future block; AR diffusion, which denoises one state at a time;
Rectified
flow~\citep{liu2022flow}, which denoises one state at a time;
DeepAR~\citep{salinas2020deepar}; SF~\citep{jiang2025streaming};
SF (perturbation), which perturbs the normalized initial state at inference;
frozen Chronos-2~\citep{ansari2025chronos}; and Chronos-2 (calibrated), our
post-hoc affine calibration of its predictive quantiles.

\paragraph{Method-specific settings.}
\begin{itemize}
  \item \textbf{Block diffusion.} A conditional one-dimensional U-Net jointly denoises a length-20 target window comprising the anchor and 19 future states. We commit 5 predicted states as a block before reconditioning. A blockwise autoregressive strategy is also used in video generation \citep{voleti2022MCVD, ho2022video}. We use 100 training noise levels, $x_0$ prediction, and DDIM~\citep{song2020denoising} with 20 denoising steps.

  \item \textbf{AR diffusion.} Following TimeGrad's autoregressive forecasting formulation~\citep{rasul2021timegrad}, we denoise one future state at a time. Our implementation uses a four-layer, four-head causal Transformer \citep{peebles2023scalable} of width 128 with diffusion $v$-prediction. We use 100 training noise levels and DDIM with 20 denoising steps per predicted state.

  \item \textbf{Rectified flow}~\citep{liu2022flow} uses a four-layer, four-head transformer with width
  128, as in AR diffusion. It autoregressively generates one future state at a time using velocity
  prediction and 10 denoising steps per predicted state.
  
  \item \textbf{DeepAR}~\citep{salinas2020deepar} is an autoregressive probabilistic forecasting model based on recurrent neural networks. We use a two-layer LSTM with hidden width 128 and a
  diagonal Gaussian output distribution with minimum scale $10^{-4}$.

  \item \textbf{SF and SF (perturbation).} SF~\citep{jiang2025streaming} uses the thin-tube path with
  $\sigma_0=0.4$ and $k=10$ and uses a U-Net backbone following the original paper and similar to block diffusion, and advances ten states before reconditioning,
  except on stochastic Lorenz where it advances one state.
  The perturbation variant modifies the normalized initial state once, at the
  start of the rollout.  We select the standard deviation of the perturbation noise  between $0.1$ and $0.01$ depending on the performance.

  \item \textbf{Chronos-2.} Chronos-2 is a pretrained time-series foundation
  model that directly returns predictive quantiles over a requested forecast
  horizon~\citep{ansari2025chronos}.  We keep its parameters frozen and use
  its full-horizon quantile forecasts.  Chronos-2 (calibrated) is our post-hoc
  variant, not part of the standard Chronos-2 method.

  \item \textbf{VSF.} VSF uses the same conditional one-dimensional U-Net as
  SF and adds a four-dimensional trajectory-level latent at the network
  bottleneck.  A two-layer MLP maps this latent to the bottleneck feature
  width.  After the middle residual blocks and before upsampling, we fuse the
  latent feature with the time- and history-conditioned bottleneck activation
  using a weighted sum.  We fix the latent
  fusion using $90\%$ latent features and $10\%$ bottleneck features, while all
  skip connections and the original SF conditioning remain unchanged.  We
  also fix the $\alpha=1.999$ and $\beta_{\mathrm{KL}}=2\times10^{-3}$ for every forecasting dataset.
  We follow SF's tube-based training construction for double pendulum and R\"ossler. We observe Lorenz dynamics
  are sensitive to injected state noise; hence we disable tube perturbations.
\end{itemize}

\paragraph{Chronos-2 calibration.}
Suppressing the context index, let $Q_{t,d}(\tau)$ be the raw Chronos-2
$\tau$-quantile at forecast time $t$ for state coordinate $d$, and let
$\mu_{t,d}:=\int_0^1Q_{t,d}(u)\,du$ be its predictive mean.  We apply the
per-coordinate transformation
\begin{equation}
  Q^{\mathrm{cal}}_{t,d}(\tau)
  =a_d+b_d\mu_{t,d}
   +c_d\bigl(Q_{t,d}(\tau)-\mu_{t,d}\bigr),
  \qquad b_d,c_d>0.
  \label{eq:chronos2-affine-calibration}
\end{equation}
Thus, $a_d$ shifts the forecast, $b_d$ rescales its mean, and $c_d$ rescales
its spread around the mean.  We compute $\mu_{t,d}$ by trapezoidal integration
over the returned quantile grid, using flat tails at 0 and 1.

\paragraph{Experimental details.}
We use 70\%/15\%/15\% train/validation/test splits over trajectories for
deterministic data and over initial conditions for parameter-mixture data.
We normalize each coordinate to $[-1,1]$ using training-set extrema and turn
the 100 trajectories per initial condition into individual training examples
only after splitting.  Learned methods
use AdamW with learning rate $10^{-4}$, zero weight decay,
a 200-step warmup followed by cosine decay, and exponential-moving-average
weights with decay $0.75$.  We train for 90 epochs on deterministic datasets
and 18 epochs on parameter-mixture datasets.  Results are the mean and
population standard deviation over three seeds.   We run all training and evaluation jobs on NVIDIA A100 GPUs. At test time we
generate 100 trajectories per held-out context and evaluate 1,000-step
rollouts; the additional stochastic-Lorenz experiment uses 2,500 steps.

\subsection{World Models}
\label{sec:world-model-setup}

\paragraph{Baseline models.}
We train the SF and VSF predictors using the JEPA-based LeWM representation
from the corresponding task-specific pretrained checkpoint~\citep{maes2026leworldmodel}.
We freeze LeWM's visual encoder, projection head, and action encoder, and compare
against its pretrained autoregressive (AR) predictor.  All
three predictors operate in the same 192-dimensional latent space and use the
same CEM planner, evaluation tasks, and planning budget.

\paragraph{Method-specific settings.}
\begin{itemize}
  \item \textbf{AR predictor.} We use the pretrained LeWM predictor, which
  autoregressively predicts the next JEPA latent conditioned on the action
  sequence.

  \item \textbf{SF predictor.} We replace the AR predictor with a deterministic
  physical-time velocity model and use one velocity-field evaluation for each
  imagined transition.

  \item \textbf{VSF predictor.} We condition the physical-time velocity on a
  prior latent sampled once and held fixed throughout each imagined rollout.
  This also requires one velocity-field evaluation per transition.
\end{itemize}

\paragraph{Experimental details.}

LeWM encodes each $224\times224$ RGB frame with a ViT-Tiny~\citep{dosovitskiy2020image} backbone using
$14\times14$ patches, 12 Transformer layers, three attention heads, and hidden
dimension 192. Its final-layer \texttt{[CLS]} token is passed through LeWM's
BatchNorm MLP projector to obtain the 192-dimensional JEPA
state used by every predictor. Each modeled transition spans five environment
steps. We use a 90\%/10\% training/validation split.  SF and VSF are
trained with AdamW using learning rate $10^{-4}$, zero weight decay, batch
size 4,096, and exponential-moving-average decay $0.75$.  We train VSF using \cref{eq:wm-vsf-teacher-forcing-loss} as the loss for 60
epochs on Push-T and Cube and 90 epochs on Two-Room and Reacher, and report
results over three seeds.

For planning, CEM maintains a Gaussian distribution over every
action sequence.   Each of 30 CEM updates evaluates 300 candidates and fits the mean and
diagonal scale to the best 30, starting from unit scale.  Following LeWM, we evaluate 50 goals per run, sample each goal 25
environment steps ahead, and allow a 50-step execution budget.  Predictor
RMSE is measured on CEM-selected action sequences.  For VSF, we keep $\alpha=1.999$. We set $\beta_{\mathrm{KL}}=1$ for PushT, Cube, and TwoRoom and $2\times10^{-3}$ for Reacher.

\clearpage
\section{Full ablation results}
\label{sec:full-ablation-results}

\begin{figure}[t!]
  \centering
  \includegraphics[width=\linewidth]{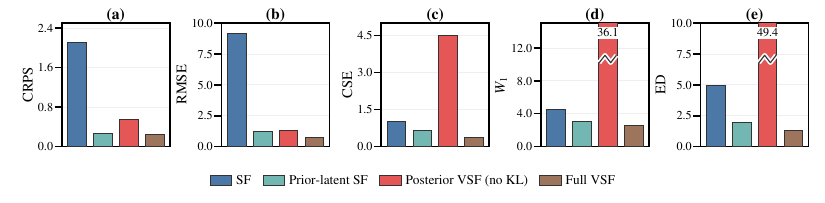}
  \caption{Latent source and prior alignment. SF, Prior-latent SF,
  Posterior no-KL, and Full VSF on (a)--(b) deterministic double pendulum and
  (c)--(e) stochastic R\"ossler. Three-seed means; breaks mark labeled
  out-of-range values.}
  \label{fig:latent-prior-alignment-ablation}
\end{figure}

Figure~\ref{fig:full-ablation-energy-score}--Figure~\ref{fig:latent-prior-alignment-ablation} report the complete ablation study
 across deterministic double pendulum and stochastic R\"ossler. Across both systems, the results support the roles of
the generalized energy loss, prior alignment, trajectory-level latent
conditioning, and rollout-wise latent consistency.

\begin{samepage}
\paragraph{Energy-score exponent $\alpha$ and MSE.}
Figure~\ref{fig:full-ablation-energy-score} shows that performance improves consistently as the energy-score exponent $\alpha$ increases, with all settings outperforming the baseline and the near-quadratic setting also surpassing MSE. This trend reflects the optimization geometry of the generalized energy score. At $\alpha=1$, it reduces to the standard energy score, whereas $1<\alpha<2$ places relatively greater emphasis on large local errors, since the gradient of a residual term $r^\alpha$ scales as $\alpha r^{\alpha-1}$. MSE remains competitive, but it lacks the pairwise term through which the energy loss directly separates multiple forecast samples.

\begin{figure}[H]
  \centering
  \includegraphics[width=\linewidth]{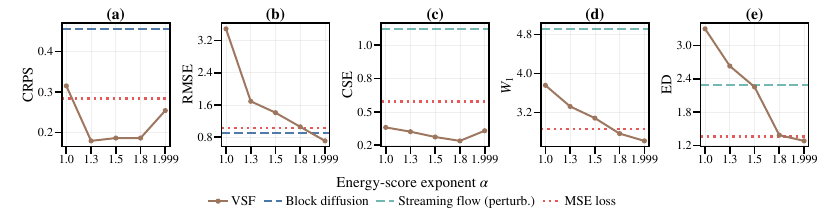}
  \caption{Energy-score exponent. $\alpha$ ablation on (a)--(b)
  deterministic double pendulum and (c)--(e) stochastic R\"ossler. Dashed
  lines show the best overall baseline for each system; dotted lines show
  MSE-trained VSF. Three-seed means.}
  \label{fig:full-ablation-energy-score}
\end{figure}
\end{samepage}

\begin{samepage}
\paragraph{KL regularization.}
Figure~\ref{fig:full-ablation-kl} shows that performance is relatively robust
across the tested KL weights and remains better overall than SF, with an
intermediate weight giving the strongest aggregate result. The KL penalty must
balance retaining target-specific branch information in the training posterior
against aligning that posterior with the prior used for forecasting. When
regularization is too weak $(\beta_{\mathrm{KL}}=1e^{-3})$, the posterior--prior gap makes prior samples less
reliable at inference.

\begin{figure}[H]
  \centering
  \includegraphics[width=\linewidth]{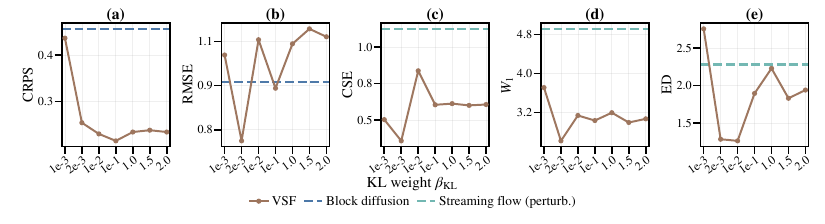}
  \caption{KL regularization. $\beta_{\mathrm{KL}}$ ablation on
  (a)--(b) deterministic double pendulum and (c)--(e) stochastic R\"ossler.
  Dashed lines show the best overall baseline for each system. Three-seed means.}
  \label{fig:full-ablation-kl}
\end{figure}
\end{samepage}

\clearpage
\begin{samepage}
\paragraph{Latent contribution.}
Figure~\ref{fig:full-ablation-latent-weight} varies the latent contribution in
bottleneck-sum fusion. With no latent contribution, the model reduces to SF
and must rely on its evolving predicted history, so rollout errors can
accumulate. Every positive latent contribution improves overall performance
relative to SF by supplying a stable trajectory-level signal that is not
corrupted by previous prediction errors. An intermediate contribution performs
best, indicating that feedback through the recurrent bottleneck remains useful.
Although performance weakens at maximal latent contribution, the model still
retains state and physical-time conditioning and therefore remains robust;
maximal latent contribution does not make the predictor independent of the
physical state.

\begin{figure}[H]
  \centering
  \includegraphics[width=\linewidth]{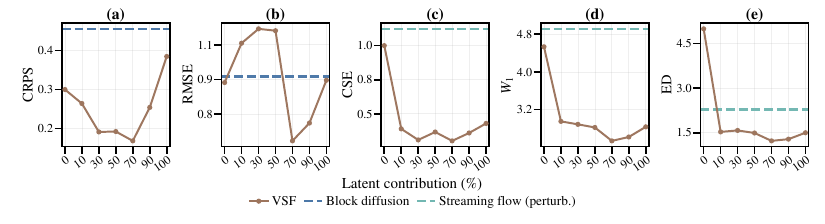}
  \caption{Latent contribution. Bottleneck latent-weight ablation
  on (a)--(b) deterministic double pendulum and (c)--(e) stochastic R\"ossler.
  Dashed lines show the best overall baseline for each system. Three-seed means.}
  \label{fig:full-ablation-latent-weight}
\end{figure}
\end{samepage}

\begin{samepage}
\paragraph{Latent resampling.}
Figure~\ref{fig:full-ablation-resampling} compares sampling one latent for the
complete rollout with drawing a new latent at every forecast step. The two
strategies perform similarly early in the forecast, but the gap widens with
horizon in favor of a fixed latent. A rollout-wise fixed $\vz$ provides a
coherent trajectory-level source of uncertainty, whereas stepwise resampling
introduces temporally inconsistent perturbations and primarily harms
long-horizon probabilistic accuracy.

\begin{figure}[H]
  \centering
  \includegraphics[width=\linewidth]{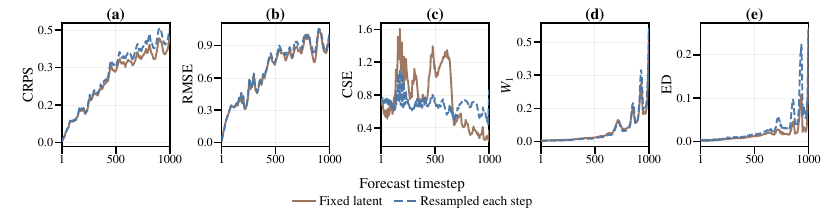}
  \caption{Latent resampling. Fixed rollout latent versus per-step
  resampling on (a)--(b) deterministic double pendulum and (c)--(e) stochastic
  R\"ossler. Three-seed means.}
  \label{fig:full-ablation-resampling}
\end{figure}
\end{samepage}

\clearpage

\paragraph{Latent-source and prior-alignment variants.}Figure~\ref{fig:latent-prior-alignment-ablation} separates three effects that
are otherwise coupled in VSF: stochastic latent conditioning,
future-conditioned variational training, and alignment of the training
posterior with the inference prior. Here
$p(\vz)=\mathcal N(\mathbf 0,\mathbf I_{d_z})$, and
$q_{\vphi}:=q_{\vphi}(\vz\mid\vu_s,\vx_s,s,\vh)$ is the
future-conditioned training posterior. Table~\ref{tab:latent-variant-definitions}
defines the four variants explicitly.

SF is the non-latent, MSE-trained reference. The three latent-conditioned
variants use the same decoder, $M$-sample generalized energy reconstruction
loss, and bottleneck latent contribution of $0.9$. Prior-latent SF is the
direct non-variational control: it uses prior samples during both training and
forecasting, with neither a posterior encoder nor a KL loss. In contrast,
Posterior VSF (no KL) remains variational because it uses the
future-conditioned posterior during training. Its prior samples at inference
are therefore unaligned with the training latents, creating a train--test
mismatch.

\begin{table}[t!]
  \centering
  \small
  \setlength{\tabcolsep}{18pt}
  \renewcommand{\arraystretch}{1.08}
  \caption{Definition of the latent-source and prior-alignment variants.}
  \begin{tabular}{@{}lcccc@{}}
    \toprule
    Variant & Training latent & \shortstack{Posterior\\encoder}
    & $\beta_{\mathrm{KL}}$ & Forecast latent \\
    \midrule
    SF & None & No & -- & None \\
    Prior-latent SF
    & $\vz^{(1:M)}\stackrel{\mathrm{iid}}{\sim}p(\vz)$
    & No & -- & $\vz\sim p(\vz)$, fixed \\
    Posterior VSF (no KL)
    & $\vz^{(1:M)}\stackrel{\mathrm{iid}}{\sim}q_{\vphi}$
    & Yes & $0$ & $\vz\sim p(\vz)$, fixed \\
    Full VSF
    & $\vz^{(1:M)}\stackrel{\mathrm{iid}}{\sim}q_{\vphi}$
    & Yes & $2\times10^{-3}$ & $\vz\sim p(\vz)$, fixed \\
    \bottomrule
  \end{tabular}

  \label{tab:latent-variant-definitions}
\end{table}

\paragraph{Does latent conditioning help?}
Prior-latent SF versus SF measures the benefit of the latent-conditioned,
multi-sample predictor without a learned posterior. Its improvement across
all  metrics shows that latent-conditioned prediction is already
useful without variational training.

\paragraph{Does variational training help?}
Full VSF has lower error than Prior-latent SF on all five metrics in Figure~\ref{fig:latent-prior-alignment-ablation}. Prior-latent SF samples training latents independently of the motion it must predict, making it harder to associate latent values with different possible futures. Full VSF instead infers latents from the observed local motion during training, helping different latent values represent different continuations. This is useful near a bifurcation, where nearly identical physical states can lead to different outcomes.

\paragraph{Why prior alignment matters.}
Full VSF versus Posterior no-KL isolates the role of the KL term when training
uses a future-conditioned posterior. Removing alignment degrades every metric,
especially the R\"ossler distributional scores. This is consistent with
train--test mismatch. Without KL
regularization, posterior codes may occupy regions or relative masses that are
not reproduced by the prior, so prior-sampled rollouts can miss or misweight
learned continuations. The gains from posterior-based training therefore
depend on aligning it with the forecasting prior.

\clearpage
\section{Rollout Error Plots}
\label{sec:rollout-error-plots}

\begin{figure}[h!]
  \centering
  \includegraphics[width=\linewidth]{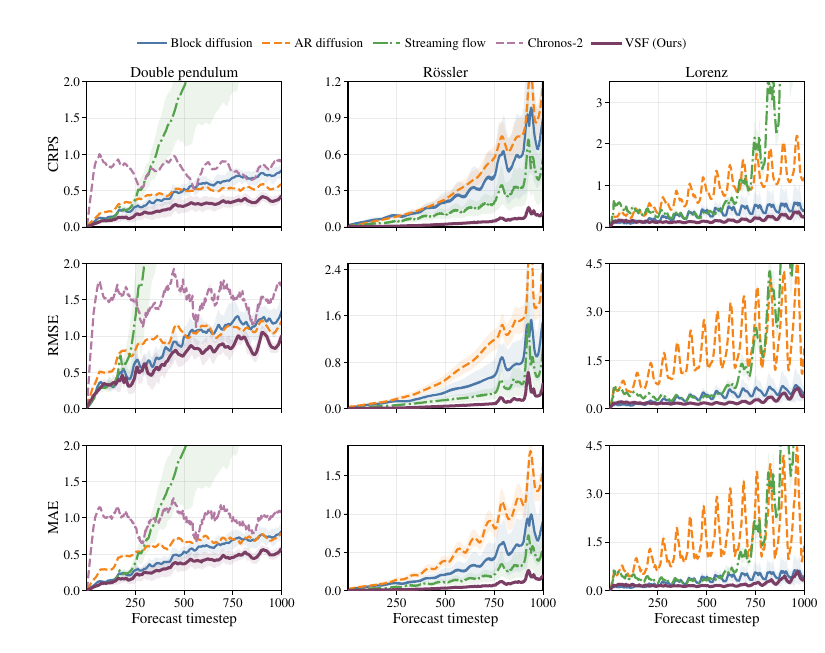}
  \caption{Per-timestep error accumulation for 1,000-step deterministic
  forecasts. Columns show double pendulum, R\"ossler, and Lorenz; rows show
  CRPS, RMSE, and MAE. Curves and shaded bands report the mean and one standard
  deviation.}
  \label{fig:rollout-errors-deterministic}
\end{figure}

\begin{figure}[h!]
  \centering
  \includegraphics[width=\linewidth]{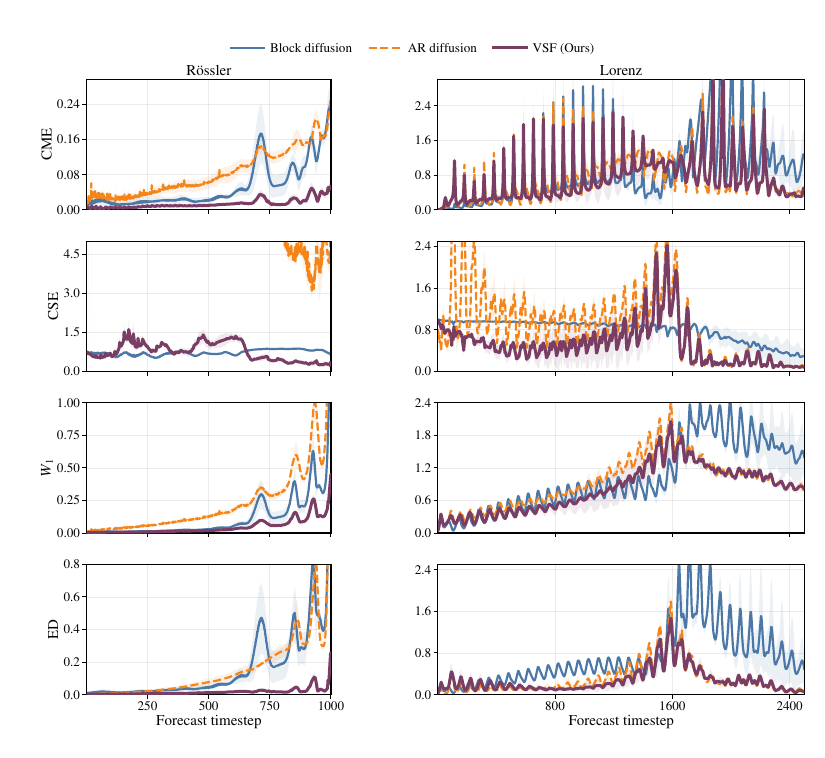}
  \caption{Per-timestep distributional error for stochastic R\"ossler over
  1,000 steps and stochastic Lorenz over 2,500 steps. Columns show the two
  systems; rows show CME, CSE, marginal $W_1$, and marginal energy distance.
  Curves and shaded bands report the mean and one standard deviation. These per-timestep scores compare marginal
  standardized state distributions. They provide a different view from the joint whole-trajectory metrics in Tables~\ref{tab:forecasting_results}
  and~\ref{tab:stochastic-lorenz-2500}.}
  \label{fig:rollout-errors-stochastic}
\end{figure}

\providecommand{\DoublePendulumInstance}{01}
\providecommand{\DoublePendulumChaoticInstance}{02}
\providecommand{\DeterministicRosslerInstance}{01}
\providecommand{\DeterministicLorenzInstance}{01}
\providecommand{\StochasticRosslerInstance}{01}
\providecommand{\StochasticLorenzInstance}{01}

\providecommand{\StochasticLorenzDrawOne}{01}
\providecommand{\StochasticLorenzDrawTwo}{08}
\providecommand{\StochasticLorenzDrawThree}{03}
\providecommand{\StochasticLorenzDrawFour}{04}
\providecommand{\StochasticLorenzDrawFive}{05}
\providecommand{\StochasticLorenzDrawSix}{06}
\providecommand{\StochasticLorenzDrawSeven}{07}

\newlength{\qualpanelwidth}
\newlength{\qualpanelheight}
\setlength{\qualpanelwidth}{0.225\linewidth}
\setlength{\qualpanelheight}{0.215\linewidth}
\newcommand{\qualplaceholder}{%
  \fbox{\parbox[c][\dimexpr\qualpanelheight-2\fboxsep-2\fboxrule\relax][c]{%
    \dimexpr\qualpanelwidth-2\fboxsep-2\fboxrule\relax}{%
    \centering\scriptsize Image\\placeholder}}%
}
\newcommand{\qualgraphic}[1]{%
  \includegraphics[
    width=\qualpanelwidth,
    height=\qualpanelheight,
    keepaspectratio
  ]{#1}%
}
\newcommand{\qualpanel}[1]{%
  \IfFileExists{#1.pdf}{%
    \qualgraphic{#1.pdf}%
  }{%
    \IfFileExists{#1.png}{\qualgraphic{#1.png}}{\qualplaceholder}%
  }%
}
\newcommand{\quallabelfont}{\normalfont\small}
\newcommand{\qualheader}[1]{{\quallabelfont\shortstack{#1}}}
\newcommand{\qualrowlabel}[1]{%
  \parbox[b][\qualpanelheight][c]{2em}{%
    \centering
    \rotatebox[origin=c]{90}{{\quallabelfont\shortstack{#1}}}%
  }%
}
\newcommand{\qualdeterministicrow}[2]{%
  \qualrowlabel{#1}
  & \qualpanel{#2_samples_gt}
  & \qualpanel{#2_samples_vsf}
  & \qualpanel{#2_mean_vsf}
  & \qualpanel{#2_samples_streaming_flow}
  & \qualpanel{#2_samples_block_diffusion} \\[0.2em]
}
\newcommand{\qualzoomrow}[2]{%
  \qualrowlabel{#1}
  & \multicolumn{5}{c}{%
    \parbox[b][\qualpanelheight][c]{%
      \dimexpr5\qualpanelwidth+8\tabcolsep\relax}{%
      \centering
      \includegraphics[
        width=\linewidth,
        height=\qualpanelheight,
        keepaspectratio
      ]{#2}%
    }%
  } \\[0.2em]
}

\newcommand{\stochasticlorenzrow}[2]{%
  {\quallabelfont #1}
  & \qualpanel{figures/lorenz_stochastic/instance_\StochasticLorenzInstance%
      _stochastic_lorenz_draw_#2_gt}
  & \qualpanel{figures/lorenz_stochastic/instance_\StochasticLorenzInstance%
      _stochastic_lorenz_draw_#2_vsf}
  & \qualpanel{figures/lorenz_stochastic/instance_\StochasticLorenzInstance%
      _stochastic_lorenz_draw_#2_streaming_flow}
  & \qualpanel{figures/lorenz_stochastic/instance_\StochasticLorenzInstance%
      _stochastic_lorenz_draw_#2_block_diffusion}
  & \qualpanel{figures/lorenz_stochastic/instance_\StochasticLorenzInstance%
      _stochastic_lorenz_draw_#2_ar_diffusion} \\[0.15em]
}

\clearpage
\section{Trajectory Forecast Visualizations}
\label{sec:qualitative-trajectory-forecasts}
We compare sampled forecast trajectories across the evaluated systems and
baselines. For the deterministic systems, we additionally show the VSF
ensemble predictive mean.

\begin{figure}[h!]
  \centering
  \begingroup
  \setlength{\qualpanelwidth}{0.18\linewidth}
  \setlength{\qualpanelheight}{0.18\linewidth}
  \setlength{\tabcolsep}{0.75pt}
  \renewcommand{\arraystretch}{1.0}
  \begin{tabular}{@{}c*{5}{c}@{}}
    & \qualheader{GT}
    & \qualheader{VSF (Ours)}
    & \qualheader{VSF ensemble}
    & \qualheader{SF}
    & \qualheader{Block diffusion} \\[0.15em]
    \qualdeterministicrow{Double \\pendulum}{figures/double_pendulum/instance_%
      \DoublePendulumInstance_deterministic_double_pendulum}
    \qualzoomrow{Double\\pendulum}{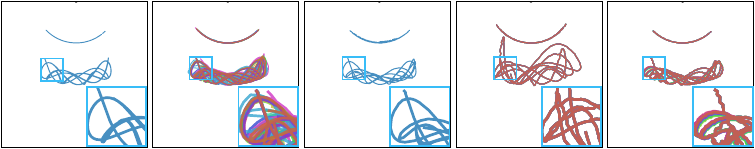}
    \qualzoomrow{R\"ossler}{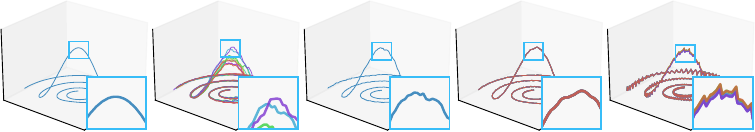}
    \qualzoomrow{Lorenz}{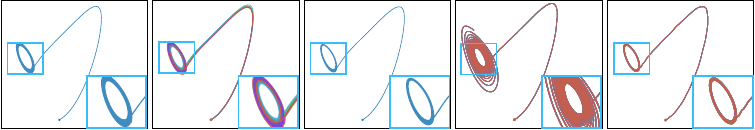}
  \end{tabular}
  \endgroup
  \caption{Qualitative forecasts on the deterministic systems. Columns show
  ground truth (GT), VSF samples, the VSF ensemble predictive mean, SF, and
  block diffusion. Sample panels contain $N=10$ trajectories; the VSF ensemble is the mean of the samples. The last three rows magnify highlighted regions of the chaotic double-pendulum, R\"ossler, and Lorenz forecasts.}
  \label{fig:qualitative-deterministic}
  \label{fig:main-chaotic-double-pendulum}
  \label{fig:qualitative-deterministic-double-pendulum-chaotic}
  \label{fig:qualitative-deterministic-rossler}
  \label{fig:qualitative-deterministic-lorenz}
\end{figure}

\begin{figure}[h!]
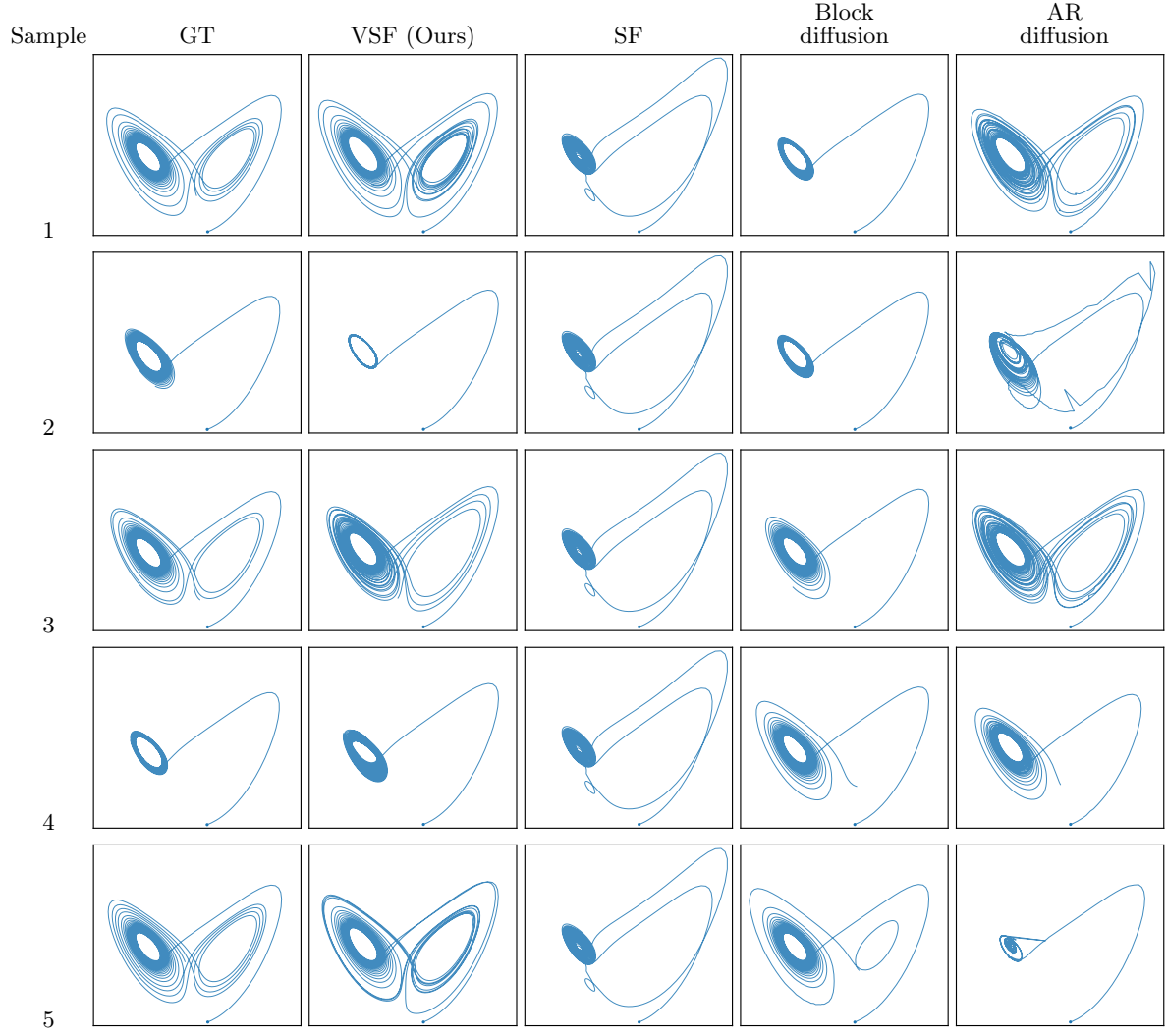

  \centering
  \begingroup
  \setlength{\qualpanelwidth}{0.18\linewidth}
  \setlength{\qualpanelheight}{0.17\linewidth}
  \setlength{\tabcolsep}{1pt}
  \renewcommand{\arraystretch}{1.05}
  \begin{tabular}{@{}c*{5}{c}@{}}
    \qualheader{Sample}
    & \qualheader{GT}
    & \qualheader{VSF (Ours)}
    & \qualheader{SF}
    & \qualheader{Block\\diffusion}
    & \qualheader{AR\\diffusion} \\
    \stochasticlorenzrow{1}{\StochasticLorenzDrawOne}
    \stochasticlorenzrow{2}{\StochasticLorenzDrawTwo}
    \stochasticlorenzrow{3}{\StochasticLorenzDrawThree}
    \stochasticlorenzrow{4}{\StochasticLorenzDrawFour}
    \stochasticlorenzrow{5}{\StochasticLorenzDrawSeven}
  \end{tabular}
  \endgroup
  \caption{Visualization samples from the ground truth and generated trajectories for the partially observed stochastic Lorenz.}
  \label{fig:qualitative-stochastic-lorenz}
\end{figure}

\clearpage
\section{World-Model Latent-Trajectory Visualizations}
\label{sec:world-model-latent-trajectory-visualizations}

\begin{figure}[H]
  \centering
  \begin{tikzpicture}
    \node[inner sep=0pt,outer sep=0pt] (arp-panel) {%
      \includegraphics[
        width=\linewidth,
        trim={0bp 0bp 0bp 20bp},
        clip
      ]{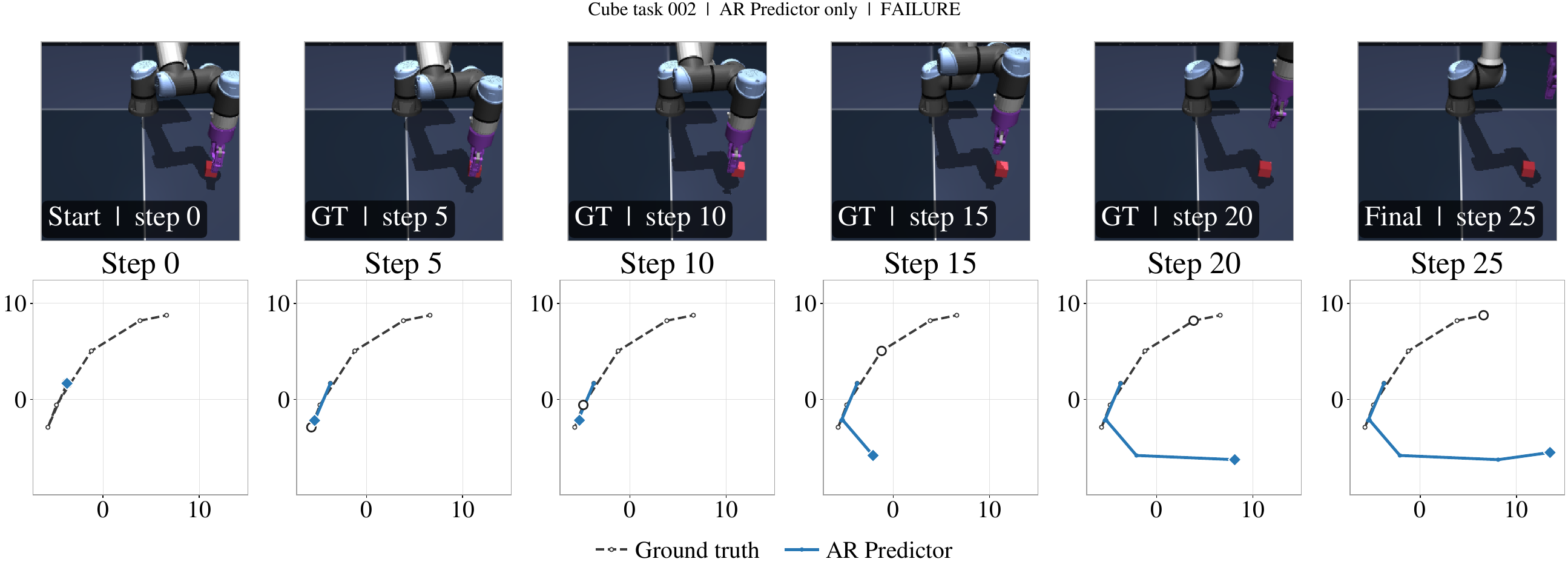}%
    };
    \node[
      anchor=south west,
      inner sep=0pt,
      outer sep=0pt,
      xshift=0pt,
      yshift=2pt,
      font=\normalfont\normalsize
    ] at (arp-panel.north west) {(a)};
  \end{tikzpicture}\par\nointerlineskip
  \begin{tikzpicture}
    \node[inner sep=0pt,outer sep=0pt] (vsf-panel) {%
      \includegraphics[
        width=\linewidth,
        trim={0bp 0bp 0bp 20bp},
        clip
      ]{figures/wm/cube_task_002_vsf_latent_trajectory.pdf}%
    };
    \node[
      anchor=south west,
      inner sep=0pt,
      outer sep=0pt,
      xshift=0pt,
      yshift=2pt,
      font=\normalfont\normalsize
    ] at (vsf-panel.north west) {(b)};
  \end{tikzpicture}
  \caption{The top row shows the ground-truth observations during rollout. The bottom row compares the predicted and ground-truth latent trajectories after joint-PCA projections. (a) The AR Predictor trajectory diverges
  and the rollout fails. (b) The VSF Predictor trajectory follows the
  ground-truth trajectory and the rollout succeeds.}
  \label{fig:wm-cube-task-002-latent-trajectories}
\end{figure}

\clearpage
\begin{figure}[H]
  \centering
  \includegraphics[
    width=\linewidth,
    trim={0bp 0bp 0bp 20bp},
    clip
  ]{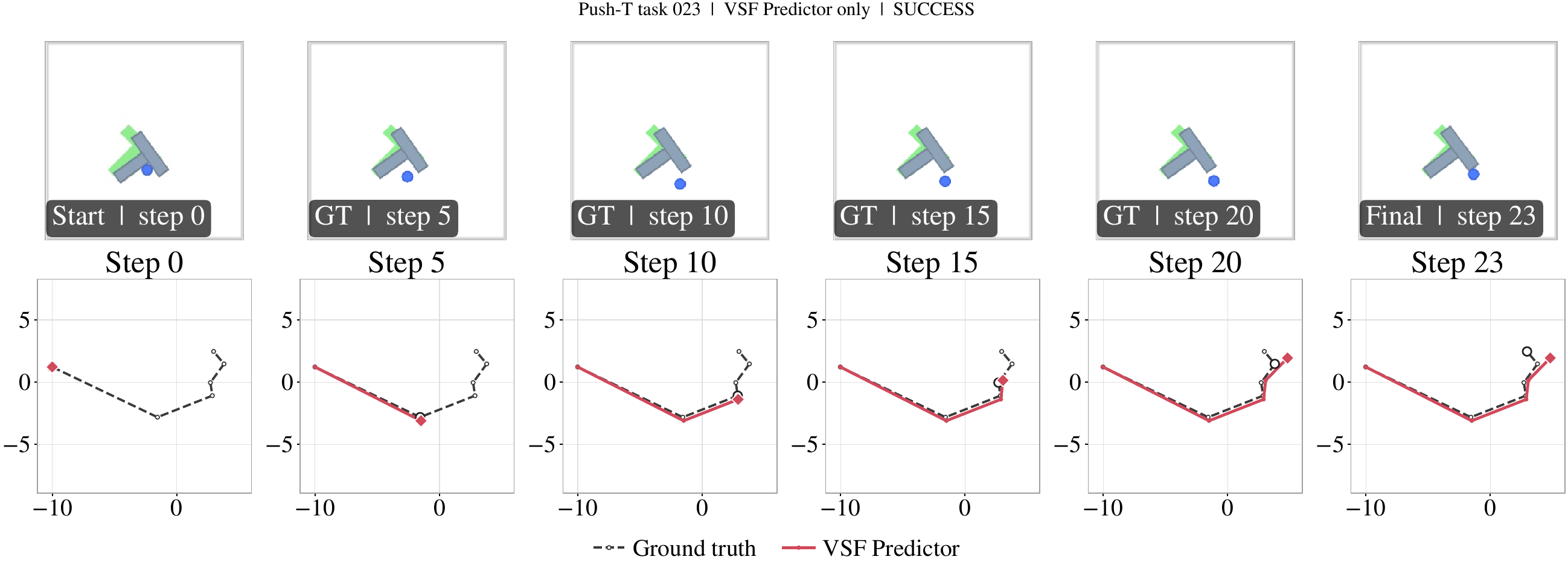}
  \caption{Qualitative PushT result. VSF action-conditioned predicted latent
  trajectory follows the ground-truth trajectory in a successful rollout.}
  \label{fig:wm-vsf-pusht-qualitative}
\end{figure}

\begin{figure}[H]
  \centering
  \includegraphics[
    width=\linewidth,
    trim={0bp 0bp 0bp 20bp},
    clip
  ]{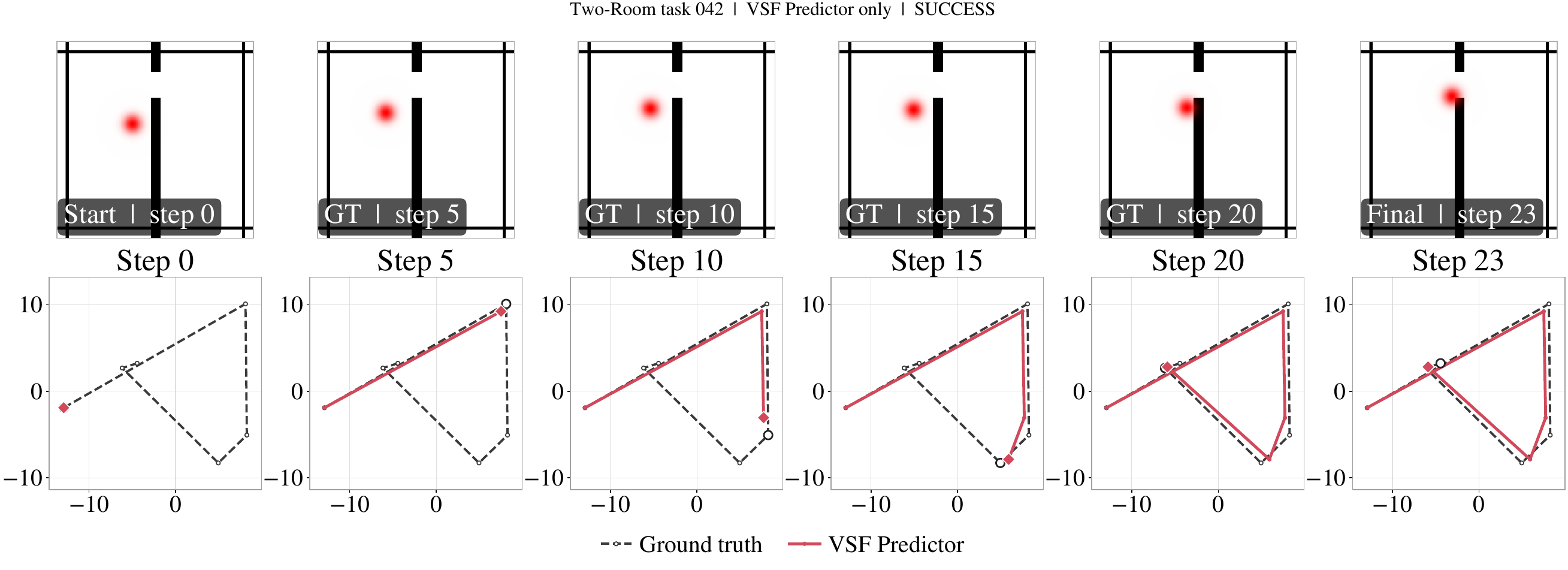}
  \caption{Qualitative TwoRoom result. VSF action-conditioned predicted latent
  trajectory follows the ground-truth trajectory in a successful rollout.}
  \label{fig:wm-vsf-reacher-qualitative}
\end{figure}

\end{document}